\documentclass[letterpaper,twocolumn,10pt]{article}
\usepackage{main}

\usepackage{tikz}
\usepackage{amsmath}

\usepackage{graphicx}

\usepackage{mathrsfs}
\usepackage{amsmath}
\usepackage[square, comma, sort&compress, numbers]{natbib}
\usepackage{algorithm}
\usepackage{algorithmicx}
\usepackage{algpseudocode}
\usepackage{booktabs}

\usepackage{enumerate}
\usepackage{multirow}
\usepackage{threeparttable}

\usepackage{amsthm}

\usepackage{color}
\usepackage[square,numbers,sort&compress]{natbib}
\usepackage{array}
\newcommand{\PreserveBackslash}[1]{\let\temp=\\#1\let\\=\temp}
\newcolumntype{C}[1]{>{\PreserveBackslash\centering}p{#1}}
\newcolumntype{R}[1]{>{\PreserveBackslash\raggedleft}p{#1}}
\newcolumntype{L}[1]{>{\PreserveBackslash\raggedright}p{#1}}

\usepackage{subfigure}
\usepackage{caption}
\usepackage{marvosym}
\usepackage{amsfonts}
\usepackage{subfigure}
\usepackage{subfigure}
\usepackage{multirow}
\usepackage[toc]{appendix}

\begin{document}

%don't want date printed
\date{}

%\title{\Large \bf FaceAdv: A Novel Method to Attack Face Recognition System in Physical World}

%\title{\Large \bf FaceAdv: A Robust Attack on Deep Learning Face Recognition in the Physical World}

\title{\Large \bf Robust Attacks on Deep Learning Face Recognition in the Physical World}

%for single author (just remove % characters)
\author{
    {\rm Meng Shen, Hao Yu, Liehuang Zhu}\\
    Beijing Institute of Technology, Beijing, China
    \and
    {\rm Ke Xu, Qi Li} \\
    Tsinghua University
    \and 
    {\rm Xiaojiang Du} \\
    Temple University
} % end author

\maketitle

%-------------------------------------------------------------------------------
\begin{abstract}

Deep neural networks (DNNs) have been increasingly used in face recognition (FR) systems.
Recent studies, however, show that DNNs are vulnerable to adversarial examples, which can potentially mislead the FR systems using DNNs in the physical world.
Existing attacks on these systems either generate perturbations working merely in the digital world, or rely on customized equipments to generate perturbations and are not robust in varying physical environments.

In this paper, we propose FaceAdv, a physical-world attack that crafts adversarial stickers to deceive FR systems.
It mainly consists of a sticker generator and a transformer,
where the former can craft several stickers with different shapes and the latter transformer aims to digitally attach stickers to human faces and provide feedbacks to the generator to improve the effectiveness of stickers.
We conduct extensive experiments to evaluate the effectiveness of FaceAdv on attacking 3 typical FR systems (i.e., ArcFace, CosFace and FaceNet).
The results show that compared with a state-of-the-art attack, FaceAdv can significantly improve success rate of both dodging and impersonating attacks.
We also conduct comprehensive evaluations to demonstrate the robustness of FaceAdv.

\end{abstract}

%-------------------------------------------------------------------------------
\section{Introduction}
%----------------------------------------------------- Introduction
Face recognition (FR) systems based on the state-of-the-art deep neural networks (DNNs) have gained utmost attention and have been widely used as a prominent biometric technique for authentication and identification in many application scenarios, such as payment authorization~\cite{wojtowicz2018face}, entry/exit management~\cite{tummon2019facial} and surveillance monitoring~\cite{rasti2016convolutional}.
%The FR system refers to facial characteristics as the unique metric to identify individuals in images.
Compared with other approaches of recognition (e.g., passwords, smart cards, voiceprint, and fingerprints), facial characteristics of an individual are relatively more difficult to be stolen, forgotten, or reproduced~\cite{zhang2020voiceprint}, and the recognition process can be carried out without the physical contact.

DNN-based FR systems, however, have been proved to be vulnerable to adversarial examples~\cite{szegedy2013intriguing}, which are derived from composing original images with \emph{perturbations} and result in an incorrect output of FR systems.
% The existing studies on adversarial examples begin in the \emph{digital} domain~\cite{szegedy2013intriguing}, where the digital adversarial examples are fabricated by tinily changing the magnitude of pixels in original images.
% The resulting perturbations are usually invisible to human eyes but can make FR systems work improperly.
% These methods cannot be directly applied in the \emph{physical} world, because potential attackers can hardly manipulate the original digital images fed into the FR systems.
Recent studies are devoted to designing effective methods for working in the \emph{physical} world~\cite{kaziakhmedov2019real,zhou2018invisible,shen2019vla}.
Due to the difference in image-forming principles between cameras and human eyes, some approaches~\cite{zhou2018invisible,shen2019vla} can avoid perturbations being observed while keeping their effectiveness.
Perturbations are projected by specialized devices to faces (e.g. a cap with LEDs~\cite{zhou2018invisible}, a projector~\cite{shen2019vla}), which spends much time to prepare and is easily discovered because of these unhidden devices.

The \emph{sticker attack} is another kind of approaches utilizing adversarial stickers to attack DNN-based recognition systems (e.g., traffic signs~\cite{eykholt2018robust,liu2019perceptual}, human faces~\cite{sharif2016accessorize,sharif2019general,komkov2019advhat}) in physical scenarios. Compared with the previously invisible attacks, the sticker attack is easy to launch but visible to human eyes. The workflow to attack FR systems can be summarized as 3 steps: 1) generating stickers in the digital world, 2) obtaining printed and tailored stickers, and 3) pasting them onto human faces~\cite{komkov2019advhat} or facial accessories~\cite{sharif2016accessorize, sharif2019general}.

% The main challenges with generating effective stickers are the \emph{sticker localization}~\cite{pautov2019adversarial} and digitally attaching stickers. Since it is obvious that we cannot paste stickers all over human faces, selecting positions to attach stickers, which is called the sticker localization, is critical to the performance of adversarial stickers. In addition, when generating digital stickers, they should be digitally attached onto face images, whose recognition results indicate the next optimization direction. Additionally, other challenges exist:
% 1) in the physical world, the distance and angle of viewing cameras and the luminance are always changing;
% 2) when attaching stickers in physical scenarios, it is difficult to paste them on exactly the same position in the digital world;
% 3) the fabrication process (e.g., printing of perturbations) is imperfect.

It is challenging to launch \emph{effective} and \emph{robust} sticker-based attacks in the physical world.
\emph{First}, as an attacker needs to paste one or multiple stickers on faces to perform dodging or impersonating attacks, it is of great importance to determine the critical positions where these stickers can be attached,
which is referred to as \emph{sticker localization}~\cite{pautov2019adversarial}.
\emph{Second}, to make the digitally designed stickers maintain their effectiveness in the physical world, the perturbed face images taken by cameras in FR systems should be efficiently and accurately simulated.
\emph{Third}, in the physical world, it is usually difficult to paste stickers exactly on the same position as design, and environmental factors (e.g., user-camera distance, brightness, and head pose) always change, resulting in severe impact on the performance of attacks.

In this paper, we propose FaceAdv, a novel attack to craft adversarial stickers to realize effective physical-world attacks. FaceAdv builds on generative adversarial networks (GANs)~\cite{gulrajani2017improved} to train a generator, i.e., a neural network that can generate adversarial stickers. When fabricated stickers are pasted on some positions (e.g., the nasal bone), we should guarantee four corners of stickers will not cover other facial organs (e.g., eyes). Thus, the generator is separated into two branches: one generates square adversarial stickers, and the other crafts masks with various shapes (e.g., circle, pentagon, and hexagon).
Once the training of the generator completes, FaceAdv can produce a large number of adversarial examples.

For tackling the first challenge, we analyze the importance of different regions of human faces and choose 5 candidate positions (i.e., two superciliary arches, the nasal bone and two nasolabial sulcus) to attach adversarial stickers. Hence, FaceAdv will generate several stickers at a time and place them on the chosen regions of human faces, which can reduce the area of each crafted sticker while keeping effectiveness.

When training the generator, the face images with stickers will be fed into the target FR system to check whether stickers successfully cheat it or not. Inspired by the R-Net~\cite{deng2019accurate} to build accurate 3D face shape from a single image, we propose a new transformer in FaceAdv to digitally stick crafted adversarial stickers on human faces for handling the second challenge.

To create robust adversarial stickers, FaceAdv draws samples (i.e., face images) from a distribution that models physical dynamics (e.g., varying distance, angles and luminance). The transformer also rotates, scales and translates stickers to simulate errors when pasting stickers on real faces in the physical world. Besides, FaceAdv introduces the total variation loss to handle the error during printing stickers.

Extensive experiments are conducted to evaluate the performance of FaceAdv. The well-known face dataset LFW~\cite{LFWTech} along with a volunteer dataset VulFace are utilized to investigate the success rate of FaceAdv and a state-of-the-art method AGNs~\cite{sharif2019general} on $3$ typical FR systems (i.e., ArcFace~\cite{deng2019arcface}, CosFace~\cite{wang2018cosface}, and FaceNet~\cite{schroff2015facenet}).
The results show that both methods could achieve high success rate in digital scenarios.
In physical scenarios, FaceAdv can significantly improve the attack success rate by $\sim50\%$ over AGNs, for both the dodging attacks and impersonating attacks.
The robustness of FaceAdv is also confirmed as its success rate is kept at a high level when environmental conditions change.

We summarize the main contributions as follows:
\begin{itemize}
  \item We propose FaceAdv to craft several adversarial stickers with different shapes at a time for more flexiblely attaching stickers to special positions (e.g., the nasal bone and the nasolabial sulcus) of human faces.
  \item We analyze the three state-of-the-art FR systems (i.e., ArcFace, CosFace, and FaceNet) to find critical regions of human face and design a transformer to digitally paste stickers onto human face.
  \item We conduct extensive experiments to demonstrate the effectiveness and robustness of stickers in both digital and physical scenarios.
\end{itemize}

We summarize the typical workflow of FR systems and existing attack algorithms in Section~\ref{sec:related_work}, and then describe the threat model in Section~\ref{sec:threat_model}. After that, we present the overview of FaceAdv in Section~\ref{sec:overview_of_faceadv} and describe the design details in Section~\ref{sec:details_of_faceadv}.
Next, we evaluate its effectiveness and robustness in Section~\ref{sec:evaluation}.
Finally, we make brief discussions in Section~\ref{sec:discussion} and conclude this paper in Section~\ref{sec:conclusion}.

\section{Background and Related Work}\label{sec:related_work}
%----------------------------------------------------- Background and Related Work

In this section, we first introduce the typical workflow of FR systems based on DNNs and describe the state-of-the-art of DNNs.
Then, we summarize the recent achievements in the digital- and physical-world attacks on FR systems.

\subsection{Background of FR Systems}\label{sec:workflow_of_frs}
%----------------------------------------------------

\begin{table*}[t]
    \caption{Summary of typical adversarial attacks on FR systems (FD: feature detection, FE: feature extraction, CF: classification)}
    \label{tab:summary}
    \centering
    \renewcommand\arraystretch{1.3}
    \scalebox{0.7}{
    \begin{tabular}{|l|l|l|p{8cm}|l|} \hline
        Domain & Attacks & Targeted Stage & Method Descriptions & Limitations\\ \hline
        \multirow{5}{*}{Digital} & Bose et al.~\cite{bose2018adversarial} & FD & \multicolumn{1}{l|}{The architecture of GANs to craft digital adversarial examples} &\multirow{5}{*}{Only working in the digital world} \\
        \cline{2-4}
        ~ & Yang et al.~\cite{yang2019design} & FD & \multicolumn{1}{l|}{Automatically designing universal patches} & ~\\
        \cline{2-4}
        ~ & $A^3GN$~\cite{song2018attacks} & FE & \multicolumn{1}{l|}{A discriminator judging between original and adversarial images} & ~\\
        \cline{2-4}
        ~ & Carofalo et al.~\cite{garofalo2018fishy} & CF & \multicolumn{1}{l|}{Deploying a poisoning attack to cheat SVM} & ~\\
        \cline{2-4}
        ~ & Dabouei et al.~\cite{dabouei2019fast} & FE \& CF & \multicolumn{1}{l|}{Manipulating the location of landmarks} & ~\\ \hline
        \multirow{7}{*}{Physical} & Kaziakhmedov et al.~\cite{kaziakhmedov2019real} & FD & \multicolumn{1}{l|}{Different face attributes printed by a white and black printer} & Only attacking the MTCNN\\
        \cline{2-5}
        ~ & IMA~\cite{zhou2018invisible} & FE & \multicolumn{1}{l|}{Projecting the perturbations on human faces using infrared} & Infrared easily filtered out with low-cost lens \\
        \cline{2-5}
        ~ & Komkov et al.~\cite{komkov2019advhat} & FE & \multicolumn{1}{l|}{An adversarial sticker attached on the hat} & Only implementing dodging attacks\\
        \cline{2-5}
        ~ & Pautov et al.~\cite{pautov2019adversarial} & FE & \multicolumn{1}{l|}{An adversarial patch pasted on different areas} & Only attacking the ArcFace system\\
        \cline{2-5}
        ~ & AGNs~\cite{sharif2019general} & FE \& CF & \multicolumn{1}{l|}{Generating adversarial stickers attached on the eyeglasses} & The colorful frames of eyeglasses look unusual\\
        \cline{2-5}
        ~ & VLA~\cite{shen2019vla} & FE \& CF & \multicolumn{1}{l|}{Hiding perturbations using the effect of PoV} & Inconvenience of launching attacks\\
        \cline{2-5}
        ~ & FaceAdv (this paper) & FE \& CF & \multicolumn{1}{l|}{Crafting stickers with different shapes attached on critical regions} & The stickers are visible to human eyes \\ \hline
    \end{tabular}
    }
\end{table*}

\begin{figure}[tb]
    \centering
    \includegraphics[width=1.0\linewidth]{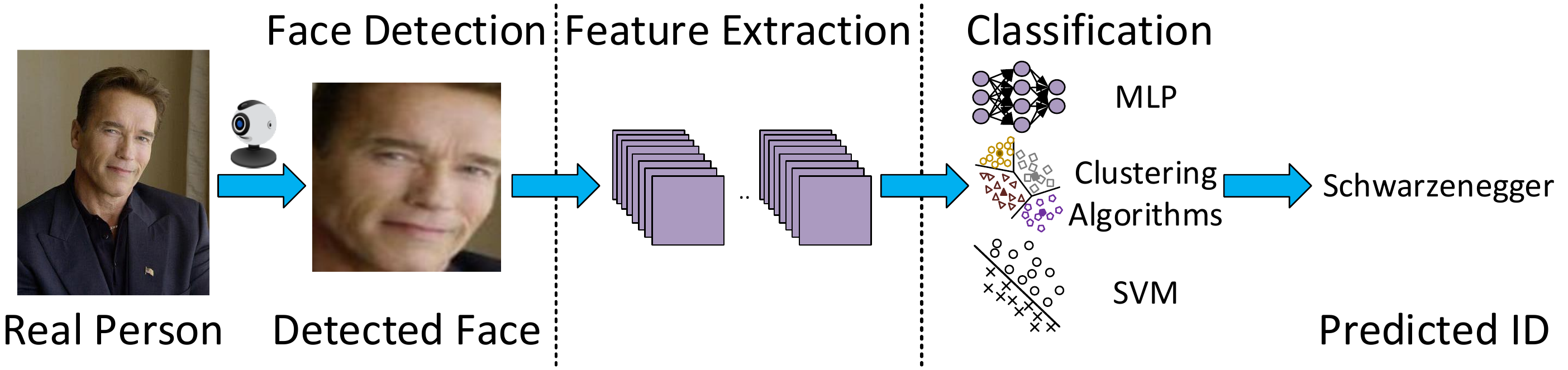}
    \caption{The typical workflow of FR systems
    %First, a face detector is used to localize faces. Second, detected faces are cut off and resized to the same scale. Third, processed faces are fed into feature extractor to geenrate embedding vectors, i.e., features corresponding to the human faces. Fourth, FR systems apply MLPs, clustering algorithms and SVMs to distinguish which person the embedding vector belongs to.
    }
    \label{fig:frs_workflow}
\end{figure}

We introduce the typical workflow of FR systems as illustrated in Fig.\ref{fig:frs_workflow}.
It generally consists of 3 steps: face detection, feature extraction, and classification~\cite{amos2016openface,garofalo2018fishy}.

\textbf{Face detection}.
The camera embedded in a FR system takes images of one's face, which are then used to localize the facial region. The purpose of this step is to determine whether human faces exist in the image or not. Once faces are detected, systems preprocess each face in the image to create the normalized and fixed-size input to the following DNNs for the feature extraction.
% Recent studies show that deep learning approaches can achieve promising performance in face detection~\cite{zhang2016joint,ming2019group}.
Since MTCNN~\cite{zhang2016joint} produces real-time and high-precision face detection results, we use it as the face detector in this paper.

\textbf{Feature Extraction}. It plays an important role in FR systems, which uses CNNs as the face embedding model to craft embedding vectors. The input is the fixed-size, frontalized face image and the output is the feature vector that describes prominent features of the face image, such as mouth, nose and eyes with their geometry distribution.
Several state-of-the-art CNNs have been proposed for the feature extraction:
\begin{itemize}
    \item FaceNet~\cite{schroff2015facenet}, which directly learns a mapping from face images to a compact Euclidean space where the distance is a measure of face similarity.
    \item CosFace~\cite{wang2018cosface}, which uses large margin cosine loss to maximize inter-class variance  (i.e., the cosine distance variance of different persons) and minimize intra-class variance (i.e., the cosine distance variance of embedding vectors of the same person) in the angular space.
    \item ArcFace~\cite{deng2019arcface}, which borrows the idea of CosFace and introduces an additive angular margin loss to obtain highly discriminative features for FR.
\end{itemize}

\textbf{Classification}.
A low-dimensional representation created by the feature extractor can be efficiently used for classification.
According to the application scenarios of FR systems, this step can be divided into a binary classification (i.e., authentication) and multi-class classification~\cite{kortli2020face}.
Several classifiers are used for either binary classification, such as SVM~\cite{garofalo2018fishy},
or multi-classification, such as Multi-Layer Perceptrons (MLPs) and K-Means.

\subsection{Summary of Attacks on FR Systems}\label{sec:summary_of_typical}
%--------------------------------------------------- Summary of Typical Attack Methods

With the wide adoption of FR systems in various scenarios, their security and safety have attracted increasing research interests.
Exiting attacks on FR systems can be roughly classified into two categories: the \emph{digital} attacks, which generate imperceptible perturbations attached directly on digital face images, and the \emph{physical} attacks, which design effective perturbations to mislead FR systems in the physical world.

Existing typical attacks on FR systems are summarized in Table \ref{tab:summary}.
These attacks target different stages of a FR system.
An adversary targeting the face detector aims to mislead the face detector not to localize the face.
Attacks on the feature extractor can reduce the distance of low-dimensional vectors of two images from different persons or enlarge the distance of two images from the same person.
Other attacks can also cheat classifiers to make incorrect decision.

\textbf{Digital attacks}. These methods directly manipulate pixel values of face images and feed the modified images into FR systems.
Except for the effectiveness, the imperceptibility of pixel-level perturbations is critical.
Thus, these algorithms always constrain the magnitude of perturbations to ensure the perturbed images are visually similar to the original images.

Bose et al.~\cite{bose2018adversarial} introduced the architecture of GANs to craft digital adversarial examples to attack the Faster R-CNN face detector.
Yang et al.~\cite{yang2019design} proposed an optimization-based approach to design universal patches, which performs well when attacking the face detector introduced by Ming et al.~\cite{ming2019group}.
Song et al.~\cite{song2018attacks} presented a specific attentional adversarial attack generative network ($A^3GN$), which leverages GANs to generate fake images to cheat the feature extractor.
%Unlike traditional GANs, they refer to the feature extractor as the third player, which can provide the cosine distance between crafted adversarial images and the image of target person to promise the fake images can successfully cheat the feature extractor.
Carofalo et al.~\cite{garofalo2018fishy} crafted adversarial images to deploy a poisoning attack against the SVM classifier. Dabouei et al.~\cite{dabouei2019fast} proposed a fast landmark manipulation method for generating adversarial faces to attack the feature extractor and classifiers.

In most cases, however, an adversary cannot directly manipulate the input images to FR systems, making these attacks not applicable in the physical world.

\textbf{Physical attacks}.
Due to spatial constraints (e.g., varying luminance and face pose), fabrication errors and resolution changes, the perturbations working well in the digital world will lose efficacy, which is referred to as \emph{perturbation loss}.
Recent studies focus on designing physical attacks to generate robust perturbations that can survive in the physical world.

In order to achieve effectiveness and imperceptibility simultaneously, several attacks utilize the difference in image-forming principles between cameras and human eyes.
Zhou et al.~\cite{zhou2018invisible} deceived the feature extractor by illuminating the subject using infrared, as infrared-based perturbations are invisible to human eyes but can be captured by cameras. However, these perturbations are easily filtered out by infrared cut-off filters that are commonly equipped in solid state cameras (e.g., CMOS).
Shen et al.~\cite{shen2019vla} proposed VLA, which leverages visible light to generate a perturbation frame and a concealing frame that are alternately projected on human faces.
Its imperceptibility relies on a phenomenon called Persistence of Vision: if the two fames changes faster than $25$Hz, human brain will mix them together and thus cannot observe perturbations. However, it requires certain kinds of equipments and is not easily conducted in real-world scenarios.

Several studies pay much attention to improving the convenience and effectiveness of adversarial perturbations rather than keeping their imperceptibility.
Kaziakhmedov et al.~\cite{kaziakhmedov2019real} proposed different face attributes printed by an ordinary white and black printer and attached them to either the medical face mask or the real face to attack MTCNN face detector.
Komkov et al.~\cite{komkov2019advhat} elaborately designed an adversarial sticker attached on the hat to cheat ArcFace feature extractor.
Unlike the two methods above, Sharif et al.~\cite{sharif2019general, sharif2016accessorize} presented white-box attacks to generate adversarial stickers attached on the eyeglasses to cheat feature extractor and MLP classifier.
They tried their best to make these adversarial stickers inconspicuous, other than imperceptible, to human eyes.
However, the colorful frames of eyeglasses still look unusual because frames are generally solid color in daily life.

%In this paper, FaceAdv takes the feature extractor and the MLP classifier as the target. Precisely speaking, adversarial stickers crafted by the generator that are pasted on the detected face image do not intentionally attack the face detector. The face images attached stickers are preprocessed and fed into the feature extractor and the MLP classifier to obtain results.

FaceAdv proposed in this paper aims at generating effective adversarial stickers to cheat the feature extractor and MLP classifiers.
%From the aforementioned approaches, physical attacks should deal with how to keep effectiveness of perturbations in uncontrolled environments.
To mitigate perturbation loss, FaceAdv applies several stickers (i.e., 3 in our implementation) attached on critical positions of human faces and introduces a series of transformations to simulate the digital-to-physical transformation process. To make the stickers relatively normal, FaceAdv presents the stickers with different shapes and the area of stickers is smaller than the stickers crafted in~\cite{komkov2019advhat}.

\section{Threat Model}\label{sec:threat_model}

In this section, we describe the threat model and design goals of the proposed FaceAdv.
%which is based on~\cite{sharif2019general, shen2019vla}.

\subsection{White-Box Assumption}\label{sec:white_box_scenario}
We follow the white-box assumption in the literature~\cite{sharif2019general},
which assumes the adversary has the full knowledge of the target FR system.
It indicates that the adversary knows the dimension of input, as well as the architecture and parameters of the feature extractor and MLP classifier.
We also assume the FR system accessed by the adversary is already well trained so that the adversary cannot manipulate the training process of the system.
Thus, the poisoning attack, which requires injecting adversarial images in the training set of FR systems, is beyond our consideration in this paper.

To train an effective generator of perturbations, the adversary can feed the images with perturbations into the target FR systems and utilize the gradients of these perturbed images for training.
This assumption is reasonable as the adversary can reconstruct the target FR systems by training a local substitute model~\cite{papernot2017practical,tang2018query}.

We select 3 state-of-the-art FR systems with different feature extractors as the target models, i.e., FaceNet~\cite{schroff2015facenet},
CosFace~\cite{wang2018cosface},
and ArcFace~\cite{deng2019arcface},
as described in Section~\ref{sec:workflow_of_frs}.
%These systems are trained in a large dataset, such as the CASIA-WebFace~\cite{yi2014learning} containing 0.49M face images from 10,575 subjects, and we should retrain a MLP classifier to distinguish our volunteers while freezing the parameters of feature extractors in ArcFace, CosFace and FaceNet to physically attack these systems, which is referred to as the \emph{fine-tuning} process.

\subsection{Attack Goals}

The goal of the adversary is to trick the FR systems to misclassify the adversarial input.
Given a FR system $\mathcal{F}_\theta(\cdot)$ with parameters $\theta$ containing the feature extractor and the classifier, and an input facial image $x$ with its ground truth label $y$ (e.g., identity), an ideal FR system can label $x$ as $y$, which is defined as $\mathcal{F}_\theta(x) = y$.
However, the adversarial image can cause the system to make an incorrect prediction.

In this paper, we consider two types of attacks.

\textbf{Dodging attacks}. The adversary aims to craft the adversarial image $x^\ast = x + \Delta_x$ with the perturbation $\Delta_x$ to mislead the classification result, which can be expressed by $\mathcal{F}_\theta(x^\ast) \neq y$.

The dodging attack enables an escaped criminal, who tries to hide himself in public video surveillance, to deceive the FR system by identifying him as someone else.

\textbf{Impersonating attacks}.
The adversary attempts to mislead the FR system by classifying the perturbed image $x^\ast= x + \Delta_x$ as a target label $y^\ast$, i.e., $\mathcal{F}_\theta(x^\ast)=y^\ast$.

In real-world applications, $y^\ast$ can be a legitimate individual with a certain authority. Such an attack enables the adversary to illegally unblock authentication.

In this paper, our primary goal is to generate adversarial examples with strong effectiveness to deceive the state-of-the-art FR systems.
The inconspicuousness of adversarial perturbations might not be necessary in certain unattended scenarios, such as unlocking a mobile phone or a car~\cite{car_recognition}, the face scan payment in unattended convenience stores~\cite{convenience_store}, and the access control of smart buildings~\cite{access_control}.
Hence, after meeting the basic requirement of effectiveness,
we try several strategies to make the attacks stealthy~\cite{sharif2019general, sharif2016accessorize}.

\section{Overview of FaceAdv}\label{sec:overview_of_faceadv}

In this section, we first describe the overview of FaceAdv, and then present a case study to show adversarial examples against target FR systems.

\begin{figure}[t]
    \centering
    \includegraphics[width=1.0\linewidth]{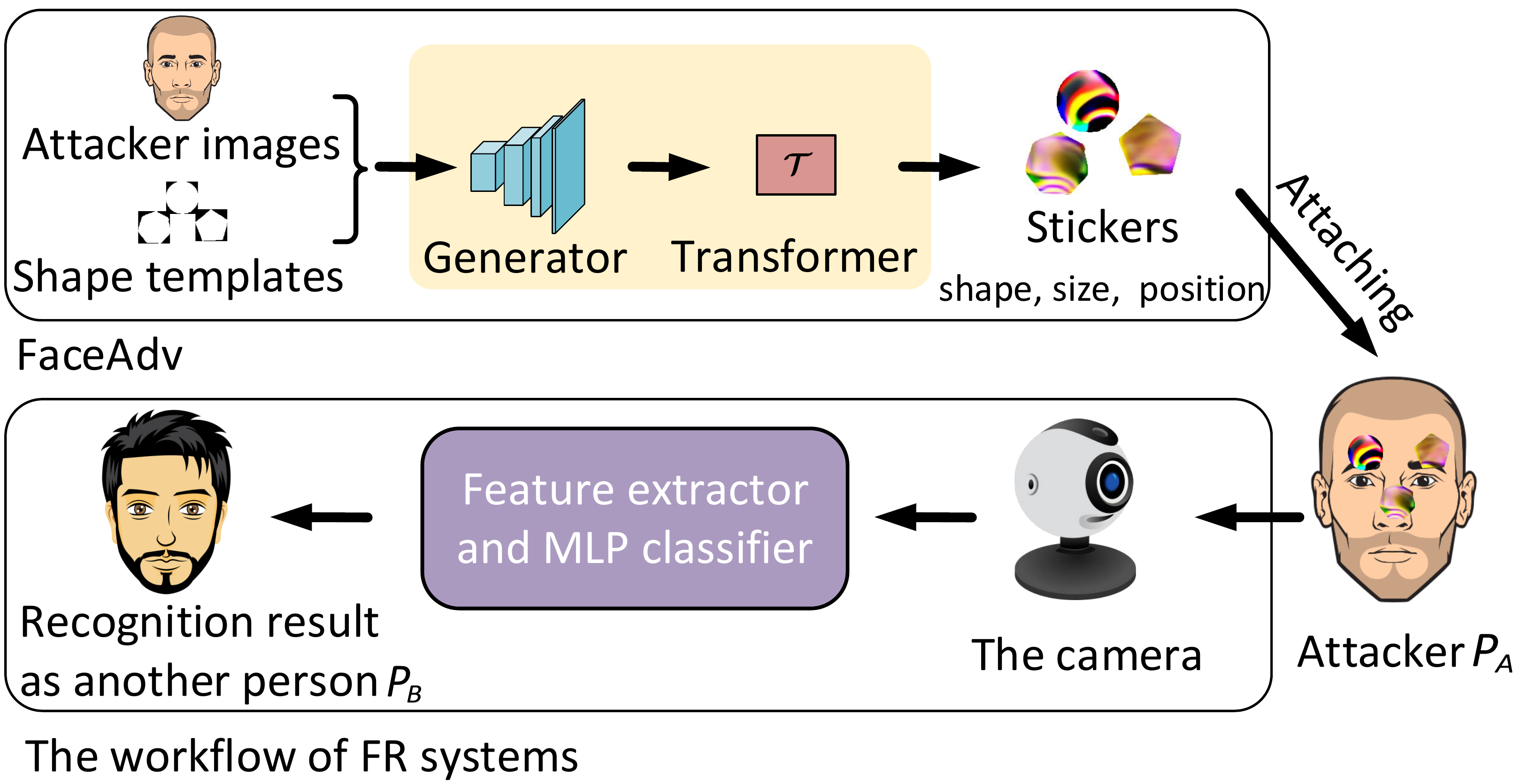}
    \caption{The physical-world attacks on FR systems using adversarial stickers crafted by FaceAdv}
    \label{fig:overview}
\end{figure}

When designing FaceAdv, we envision it will be easy to be launched in the physical world.
Therefore, we resort to adversarial stickers that can be simply stuck onto human faces, which requires no special devices such as the LEDs or projectors~\cite{zhou2018invisible,shen2019vla}.  Considering the perturbation loss from the digital to physical world, it is a challenging task to making the sticker-based adversarial examples effective and robust in varying environmental conditions.

% In this subsection, we will elaborate on the design of the transformer $\mathcal{T}$, which tackles the challenges above.
The workflow of FaceAdv is illustrated in Fig.~\ref{fig:overview}.
The attacking strategy can be explained as a sequence of the following steps:
1) taking several photos of attacker face in different environmental conditions, e.g., user-camera distance, brightness and head poses;
2) choosing desired locations and shapes of stickers;
3) training the generator of GANs with attacker face images and shape templates. During this process, $\mathcal{T}$ digitally attaches stickers to face images;
4) generating stickers including shapes, sizes and positions and printing stickers on paper;
5) standing in front of the camera and attacking FR systems.

To achieve high effectiveness, FaceAdv can craft several stickers pasted on different regions of a human face.
In general, the more stickers pasted on a face, the higher success rate FaceAdv has.
In an extreme case, the entire face is covered by crafted stickers, which seems that the attacker wears a mask.
However, the face liveness detection in FR systems can discover mask-based attacks~\cite{farrukh2020facerevelio,killiouglu2017anti}.
Thus, we should limit the area of stickers in order to pass the face liveness detection and to increase the inconspicuousness of the attack.

%However, it should be noted that FaceAdv does not compulsorily constrain the number of stickers and generates other number of stickers by altering the architecture of the generator $\mathcal{G}$ if it is necessary.
%In an extreme case, the entire face is covered by crafted stickers, which seems that the attacker wears a mask.

\begin{table}[t]
    \caption{Samples of dodging and impersonating attacks}
    \label{tab:case_study}
    \centering
    \scalebox{0.60}{
        \renewcommand\arraystretch{1.3}
        \begin{tabular}{cccccc} \hline
            \multirow{2}{*}{Attacker} & \multirow{2}{*}{Mode} & \multirow{2}{*}{Target} & \multicolumn{3}{c}{Target Model} \\
            \cline{4-6}
            ~ & ~ & ~ & ArcFace & CosFace & FaceNet \\ \hline
            \multirow{2}{*}{
                \begin{minipage}[t]{0.2\columnwidth}
                \centering
                \vspace{-20pt}
                \includegraphics[width=\linewidth]{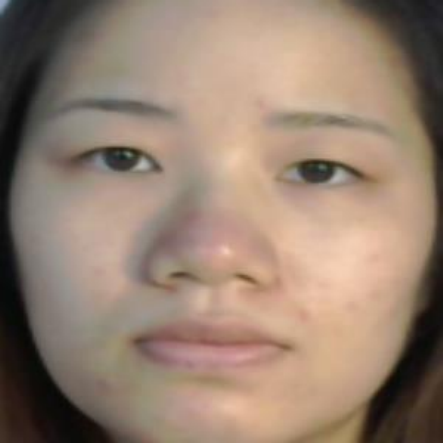}
            \end{minipage}} & \raisebox{3\height}{Dodging} &  \raisebox{3\height}{Another person} &
            \begin{minipage}[b]{0.2\columnwidth}
                \centering
                \vspace{5pt}\includegraphics[width=\linewidth]{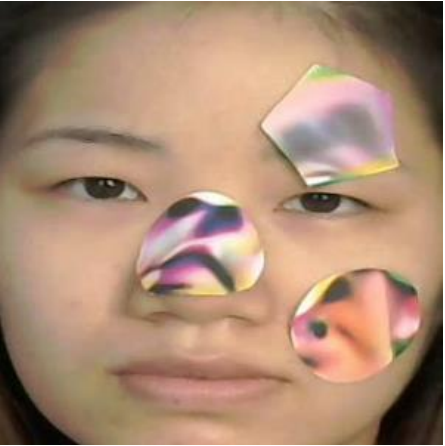}
            \end{minipage} &
            \begin{minipage}[b]{0.2\columnwidth}
                \centering
                \includegraphics[width=\linewidth]{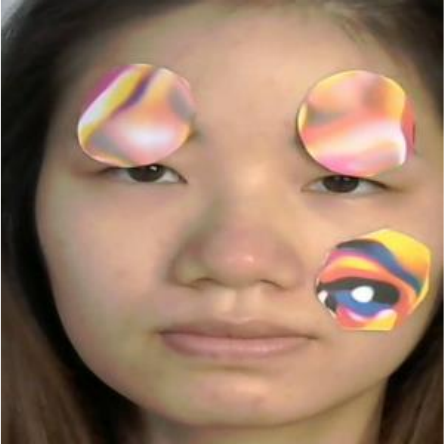}
            \end{minipage} &
            \begin{minipage}[b]{0.2\columnwidth}
                \centering
                \includegraphics[width=\linewidth]{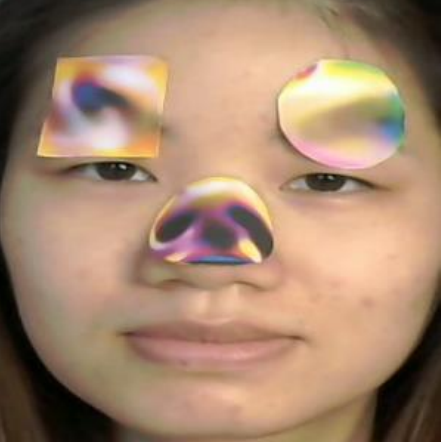}
            \end{minipage} \\
            \cline{2-6}
            ~ & \raisebox{3\height}{Impersonating} &
            \begin{minipage}[b]{0.2\columnwidth}
                \centering
                \vspace{5pt}\includegraphics[width=\linewidth]{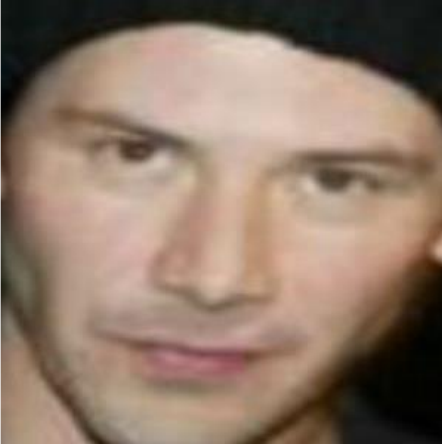}
            \end{minipage} &
            \begin{minipage}[b]{0.2\columnwidth}
                \centering
                \includegraphics[width=\linewidth]{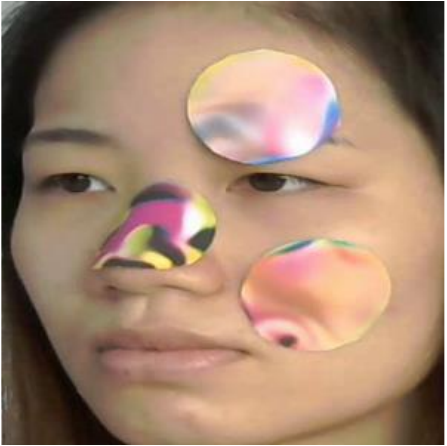}
            \end{minipage} &
            \begin{minipage}[b]{0.2\columnwidth}
                \centering
                \includegraphics[width=\linewidth]{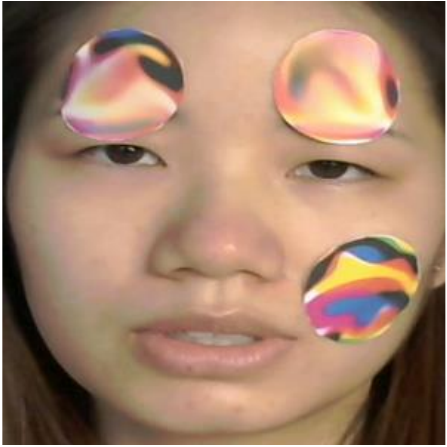}
            \end{minipage} &
            \begin{minipage}[b]{0.2\columnwidth}
                \centering
                \includegraphics[width=\linewidth]{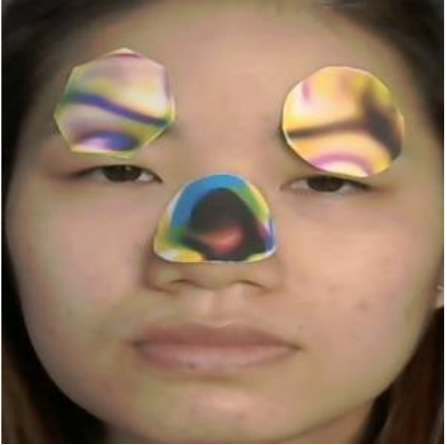}
            \end{minipage} \\ \hline
        \end{tabular}
    }
\end{table}

Next, we present an example to launch dodging and impersonating attacks on target FR systems using adversarial stickers crafted by FaceAdv, as illustrated in Table~\ref{tab:case_study}.

% {\color{blue}
% In dodging attacks, the target is random because these stickers can ensure the recognition result is not the attacker and cannot specify the recognition result. Only in impersonating attacks, the recognition result can be arranged.}

In dodging attacks, the victim is another person except for the attacker. However, in impersonating attacks, these stickers can ensure the recognition result is the attacker (i.e., Keanu Reeves).
We choose a combination of stickers for different FR systems.
When attacking the ArcFace (the fourth row and fourth column in Table~\ref{tab:case_study}), the success rate is the highest when the head of the attacker turns to right. The reason will be described in Section~\ref{sec:head_pose}.

%From the above overview of FaceAdv, the generator $\mathcal{G}$ plays such an important role that it determines the attack success rate of crafted stickers.
%In the next section, we will describe how to introduce GANs to generate stickers with different shapes and ensure the crafted stickers can meet our objectiveness in dodging attacks and impersonating attacks.
\section{Details of FaceAdv}\label{sec:details_of_faceadv}

In this section, we present design details of FaceAdv, including training the perturbation generator and digitally attaching stickers on human faces.

\begin{figure*}
    \centering
    \includegraphics[width=0.8\linewidth]{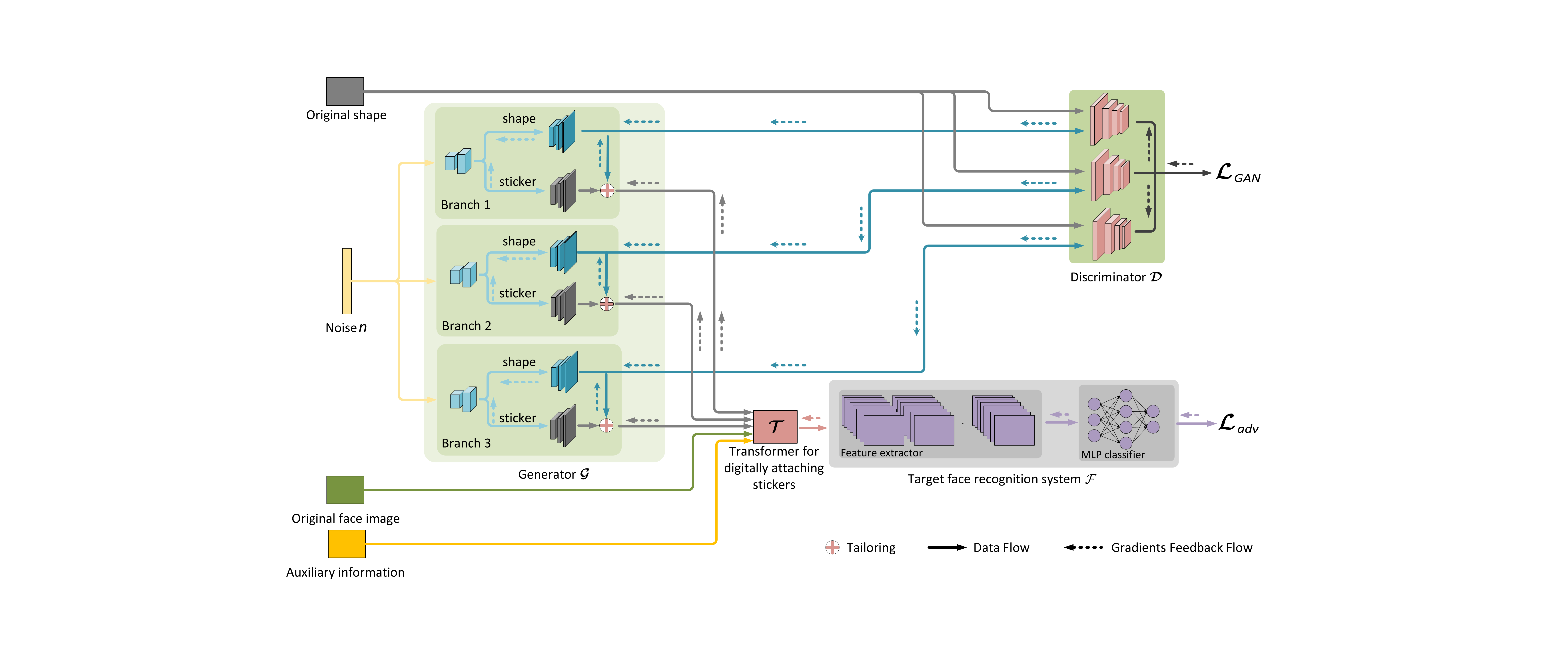}
    \caption{The FaceAdv architecture. The generator $\mathcal{G}$ is divided into three branches for crafting three stickers, and the discriminator $\mathcal{D}$ is utilized to judge whether the input shape is created by $\mathcal{G}$ or not. The stickers are fabricated by tailoring the original square rectangular according to the fake shapes. Then, the tailored stickers are digitally applied onto human faces using the original images and auxiliary information, which are finally fed into the FR system to obtain the recognition result.}
    \label{fig:gans}
\end{figure*}

\subsection{FaceAdv Architecture}\label{sec:faceadv_architecture}

\begin{table}[t]
    \caption{Notations used in this paper}
    \label{tab:notations}
    \small
    \centering
    \renewcommand\arraystretch{1.3}
    \scalebox{0.9}{
        \begin{tabular}{|l|p{7cm}|} \hline
            Notation & Discription \\ \hline
            $x_A$ & An face image of user A \\ \hline
            $P_A$ & The label of user A \\ \hline
            $\gamma_{i}$ & The parameters of the illumination model, $i\in[1, 27]$ \\ \hline
            $n$ & Noise sampled from the normal distribution \\ \hline
            $\mathcal{G}$ & The generator to craft stickers with different shapes \\ \hline
            $\mathcal{D}$ &  \multicolumn{1}{m{7cm}|}{The discriminator to discriminates between crafted shapes and real shapes} \\ \hline
            $\mathcal{F}$ & The target face recognition system \\ \hline
            $\mathcal{T}$ & The transformer to digitally attach stickers to faces \\ \hline
            $\mathcal{L}_{GAN}$ & The adversarial loss of GANs \\ \hline
            $\mathcal{L}_{adv}$ & The loss for fooling the face recognition system \\ \hline
        \end{tabular}
    }
\end{table}

As mentioned above, adversarial stickers crafted by FaceAdv should be pasted on human faces to attack FR systems, which raises three challenging problems.
\emph{First}, since traditional stickers are usually in the form of rectangle and square~\cite{pautov2019adversarial,komkov2019advhat}, four corners of a sticker may cover five sense organs when it is attached to specific locations (e.g., the nasal bone).
\emph{Second}, FaceAdv can generate several stickers to cheat systems at a time, so it is crucial to determine the positions where these stickers can be pasted.
Pautov et al.~\cite{pautov2019adversarial} place crafted adversarial stickers onto difference positions of human face (i.e., eyeglasses, forehead and nose) to attack the FR system and find positions of stickers have a dramatic influence on the effectiveness of stickers.
\emph{Third}, how to efficiently and accurately obtain the camera-perceived face images when stickers are pasted on real human faces.
In this paper, we design FaceAdv architecture to craft adversarial stickers, which is composed of the sticker generator $\mathcal{G}$ and a transformer $\mathcal{T}$ to separately takle the first challenge and the other two challenges above.

% GANs, which casts generative modeling as the two-player game: a generator $\mathcal{G}$ and a discriminator $\mathcal{D}$, are a powerful class of generative models.
% The generator network maps a source of noise to the space of real shape templates. The discriminator network receives either the fake shape crafted by the generator or the true shape from shape templates and distinguish between the two.
% Thus, with the help of the discriminator $\mathcal{D}$, the generator $\mathcal{G}$ can generate the shape in shape templates.

The sticker generator $\mathcal{G}$ can generate adversarial stickers and shapes with four corners cut off by GANs. GANs, which casts generative modeling as the two-player game: a generator $\mathcal{G}$ and a discriminator $\mathcal{D}$, are a powerful class of generative models. The generator network maps a source of noise to the space of real shape templates. The discriminator receives either the fake shape crafted by the generator or the true shape from shape templates and distinguish between the two.

FaceAdv utilizes crafted shapes to tailor stickers fabricated by the generator and the transformer $\mathcal{T}$ to digitally attach stickers to face images. Then, face images with stickers are fed into FR system, and the generator will update parameters according to recognition results. In the next two subsections, we will elaborately describe the sticker generator $\mathcal{G}$ and the transformer $\mathcal{T}$.

\subsection{Sticker Generator}\label{sec:sticker_generator}

Inspired by producing adversarial stickers placed on eyeglasses to cheat FR systems in the real world~\cite{sharif2019general}, we introduce the architecture of GANs to generate stickers. However, the difference is that we constrain the shape rather than the content of stickers using GANs.
% {\color{blue}Since the stickers created by the generator look like the real pattern, if the generator constrains the content, the real pattern can also attack the FR system if the crafted stickers has high fidelity, which means that researchers could takes a long time to find appropriate training dataset.}

Since adversarial stickers crafted by previous approaches are always square or rectangular in shape, stickers with different shapes can also lower vigilance of people nearby.
In this work, we employ the architecture of GANs to craft adversarial stickers with different shapes in Fig.~\ref{fig:gans}.
Unlike the traditional GANs, there are three players in FaceAdv.

$\mathcal{G}$ consists of three branches, each of which is utilized to generate a shape mask and an adversarial sticker. The shape mask is a binary image that only has two colors (black and white), which is used to tailor original square stickers so that the shape of cropped stickers is the same as the shape in training datasets (specifically, the shape template). The goal of the discriminator $\mathcal{D}$ is to distinguish the shapes crafted by $\mathcal{G}$ for constraining that the crafted shapes are all in original shapes. The generator $\mathcal{G}$ and the discriminator $\mathcal{D}$ are normal components of GANs. In addition, there is a third player (i.e., $\mathcal{F}$) to obtain recognition results, whose parameters cannot be changed in the process of training $\mathcal{G}$ and $\mathcal{D}$. The aim of $\mathcal{F}$ is to inform $\mathcal{G}$ the effectiveness of crafted stickers so that it can adjust the optimization direction in time.

In this paper, the proposed FaceAdv can generate three different stickers at a time (i.e., branches $1-3$). The more details of the architecture are shown in Appendix~\ref{sec:architecture_of_gans}.
However, FaceAdv has no restriction on the number of adversarial stickers. When changing the number of stickers, what to be modified is adding branches of $\mathcal{G}$ and changing the corresponding number of branches in $\mathcal{D}$ and also altering sticker positions of the transformer $\mathcal{T}$ that realizes digitally attaching stickers on human faces.
Based on our preliminary experiments, we select $3$ stickers to attack FR systems, because it can achieve better balance between effectiveness and convenience.

Since FaceAdv tailors some parts of square stickers to form stickers with different shapes, it is important to design a reasonable mechanism to assure the cut part cannot largely damage the effectiveness of adversarial stickers. In Fig.~\ref{fig:gans}, the gradients from $\mathcal{D}$ can place restrictions on the shapes of stickers.
Besides, the gradients from $\mathcal{F}$ are applied to change the parameters of $\mathcal{G}$ containing parts for crafting shapes and stickers. Apparently, the gradients flowing into the part of creating stickers can alter the content of stickers, and the gradients flowing into that of producing shapes can also change the shapes of stickers. In this way, if FaceAdv tailors some important regions, $L_{adv}$ that indicates the effectiveness of adversarial stickers will be larger and $\mathcal{G}$ will realize this problem and update parameters.

To improve the robustness of crafted stickers, when training $\mathcal{G}$, original face images are composed of multiple images captured in different conditions so that stickers can work in physical scenarios.

There are two loss functions: $\mathcal{L}_{GAN}$ and $\mathcal{L}_{adv}$,
where $\mathcal{L}_{GAN}$ is utilized to train the shape branch of $\mathcal{G}$ and $\mathcal{D}$, and $\mathcal{L}_{adv}$ is applied to optimize the sticker branch of $\mathcal{G}$. They will be presented in Section~\ref{sec:loss_designment}.

% is used to measure the performance of adversarial stickers. However, $\mathcal{L}_{GAN}$ is more complex, and we will describe it in Section~\ref{sec:loss_designment}.

There is another key part of FaceAdv to be elaborated: the transformer $\mathcal{T}$, which connects the generator $\mathcal{G}$ and the target FR system $\mathcal{F}$. As mentioned previously, FaceAdv applies 3D face shape to digitally paste adversarial stickers onto human faces and the process of attaching stickers should be differentiable so that the gradients from $\mathcal{F}$ can smoothly flow into $\mathcal{G}$. And the auxiliary information is used in $\mathcal{T}$, which includes the 3D face shape and the parameters of illumination models as described in next subsection.

\subsection{Digital Transformation of Physical Sticker}\label{sec:digital_transformation_of_physical_sticker}

As mentioned above, the transformer $\mathcal{T}$ digitally attaches adversarial stickers to crucial regions of face images.
In this subsection, we will elaborate on the design of the transformer.

\subsubsection{Sticker Localization}\label{sec:sticker_localization}

The locations of stickers on real faces can largely affect the attack effectiveness.
To maximize their effects, a natural idea is to place the stickers on the positions where the target FR system extracts discriminative features of human faces.
%Consequently, sticker localization is converted to localize the important regions in the image for FR systems predicting the identity.

To investigate regions of human faces where each target FR system extract features,
we leverage Guided Grad-CAM~\cite{selvaraju2017grad} to analyze effective regions of the three FR systems, as illustrated in Fig.~\ref{tab:grad_cam}.
The Guided Grad-CAM uses the gradient information flowing into the last convolutional layer of the feature extractor to analyze the importance of each region in the image for a decision of interest.
More details about Guided Grad-CAM are described in Appendix~\ref{sec:guided_grad_cam}.

The localization maps of 4 individuals\footnote{They are Arnold Schwarzenegger, George W Bush, Roh Moo-hyun and Vladimir Putin.} in Fig.~\ref{tab:grad_cam} show that these FR systems invariably focus on regions near to the five sense organs (e.g., eyes, nose, and mouth) to make a decision.
In ArcFace, the regions to extract features are not always near to five sense organs, e.g., temples of Schwarzenegger in Fig.~\ref{tab:grad_cam}. And also, in some cases, CosFace and FaceNet extract valuable features near to the nasolabial sulcus.

Due to face liveness detection and the convenience of tailoring adversarial stickers, it is important to balance the number of stickers and the area of each sticker.
When fixing the total area of perturbed regions on human faces, if the number of stickers is larger, the area of each sticker will be smaller, but the time to tailor stickers will be longer. In this paper, we choose $3$ adversarial stickers to achieve a better balance. %because in this number of stickers, the time to print, cut and attach stickers will be less than 5 minutes.

\begin{table}[t]
    \caption{The localization maps crafted by Guided Grad-CAM}
    \label{tab:grad_cam}
    \centering
    \scalebox{0.52}{
        \renewcommand\arraystretch{1.3}
        \begin{tabular}{ccccc} \hline
            \multirow{2}{*}{Original Images} & \multicolumn{3}{c}{Model} & \multirow{2}{*}{3D Faces} \\
            \cline{2-4}
            ~ & ArcFace & CosFace & FaceNet & ~\\ \hline
            % Schwarzenegger
            \begin{minipage}[b]{0.2\columnwidth}
                \centering
                \vspace{5pt}\includegraphics[width=\linewidth]{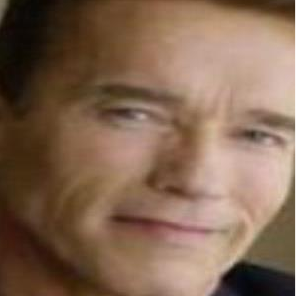}
            \end{minipage} &
            \begin{minipage}[b]{0.2\columnwidth}
                \centering
                \includegraphics[width=\linewidth]{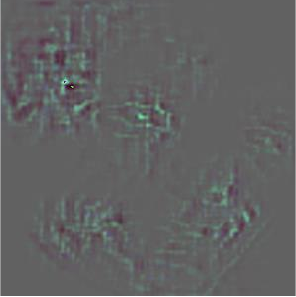}
            \end{minipage} &
            \begin{minipage}[b]{0.2\columnwidth}
                \centering
                \includegraphics[width=\linewidth]{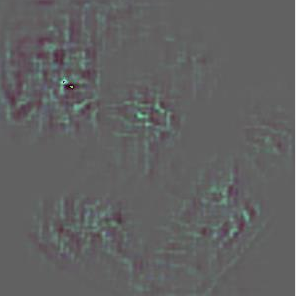}
            \end{minipage} &
            \begin{minipage}[b]{0.2\columnwidth}
                \centering
                \includegraphics[width=\linewidth]{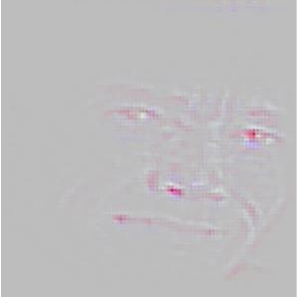}
            \end{minipage} &
            \begin{minipage}[b]{0.18\columnwidth}
                \centering
                \includegraphics[width=\linewidth]{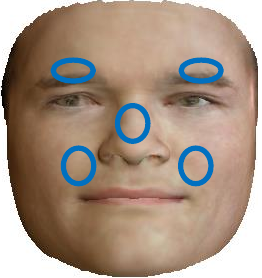}
            \end{minipage} \\ \hline
            % Bush
            \begin{minipage}[b]{0.2\columnwidth}
                \centering
                \vspace{5pt}\includegraphics[width=\linewidth]{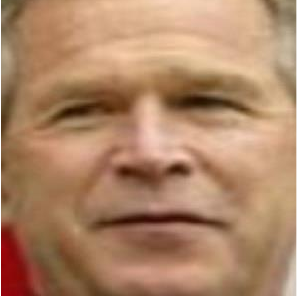}
            \end{minipage} &
            \begin{minipage}[b]{0.2\columnwidth}
                \centering
                \includegraphics[width=\linewidth]{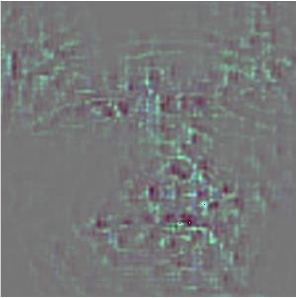}
            \end{minipage} &
            \begin{minipage}[b]{0.2\columnwidth}
                \centering
                \includegraphics[width=\linewidth]{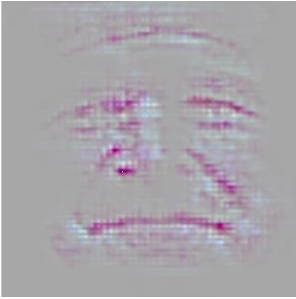}
            \end{minipage} &
            \begin{minipage}[b]{0.2\columnwidth}
                \centering
                \includegraphics[width=\linewidth]{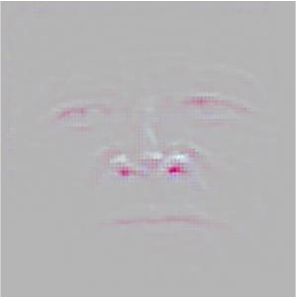}
            \end{minipage} &
            \begin{minipage}[b]{0.18\columnwidth}
                \centering
                \includegraphics[width=\linewidth]{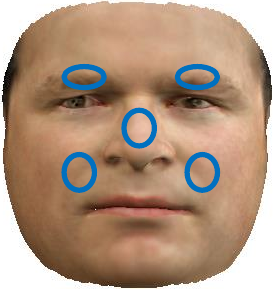}
            \end{minipage} \\ \hline
            % Moo-Hyun
            \begin{minipage}[b]{0.2\columnwidth}
                \centering
                \vspace{5pt}\includegraphics[width=\linewidth]{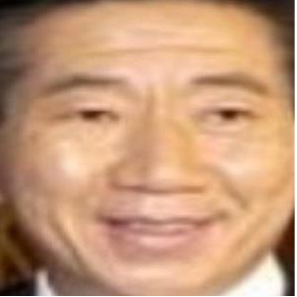}
            \end{minipage} &
            \begin{minipage}[b]{0.2\columnwidth}
                \centering
                \includegraphics[width=\linewidth]{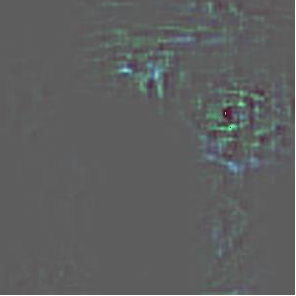}
            \end{minipage} &
            \begin{minipage}[b]{0.2\columnwidth}
                \centering
                \includegraphics[width=\linewidth]{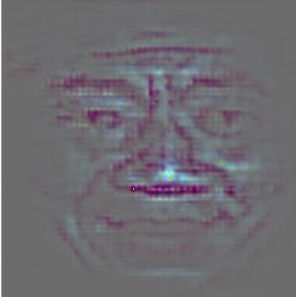}
            \end{minipage} &
            \begin{minipage}[b]{0.2\columnwidth}
                \centering
                \includegraphics[width=\linewidth]{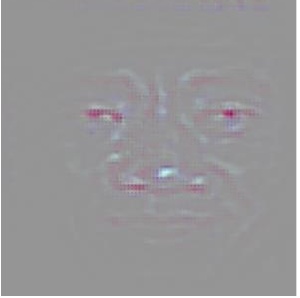}
            \end{minipage} &
            \begin{minipage}[b]{0.18\columnwidth}
                \centering
                \includegraphics[width=\linewidth]{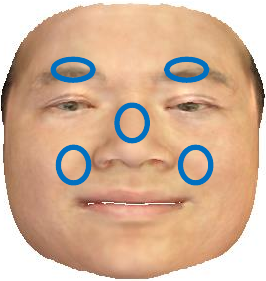}
            \end{minipage} \\ \hline
            % Putin
            \begin{minipage}[b]{0.2\columnwidth}
                \centering
                \vspace{5pt}\includegraphics[width=\linewidth]{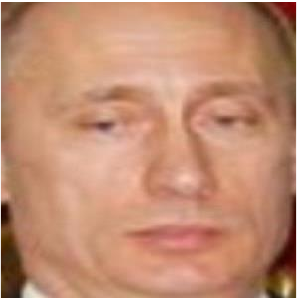}
            \end{minipage} &
            \begin{minipage}[b]{0.2\columnwidth}
                \centering
                \includegraphics[width=\linewidth]{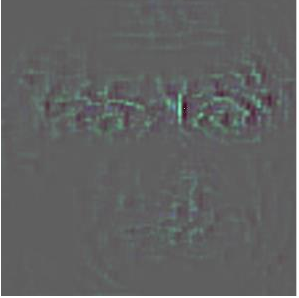}
            \end{minipage} &
            \begin{minipage}[b]{0.2\columnwidth}
                \centering
                \includegraphics[width=\linewidth]{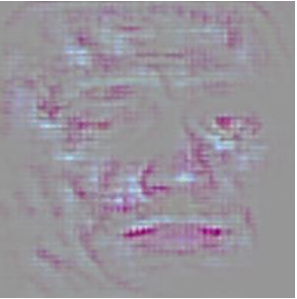}
            \end{minipage} &
            \begin{minipage}[b]{0.2\columnwidth}
                \centering
                \includegraphics[width=\linewidth]{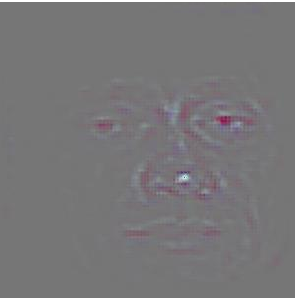}
            \end{minipage} &
            \begin{minipage}[b]{0.18\columnwidth}
                \centering
                \includegraphics[width=\linewidth]{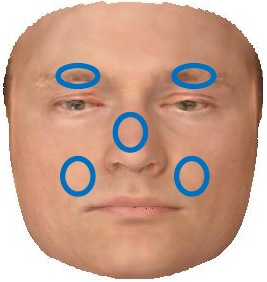}
            \end{minipage} \\ \hline
        \end{tabular}
    }
\end{table}

The stickers crafted by FaceAdv cannot directly cover five sense organs, so FaceAdv tends to attach the stickers in regions near to five sense organs.
Based on the observations above, we select $5$ regions (i.e., two superciliary arches, two nasolabial sulcus, and the nasal bone) as the candidate regions for pasting adversarial stickers generated by FaceAdv, as marked by the blue circles in the last column of Table~\ref{tab:grad_cam}.

Existing studies have also shown the effectiveness of adversarial stickers pasted on these regions, e.g., two superciliary arches are selected by~\cite{komkov2019advhat,sharif2019general}, the nasal bone is chosen by~\cite{pautov2019adversarial}.
We will describe the selection of final regions to attach stickers for each target FR system in Section~\ref{sec:locations_of_stickers}.

\subsubsection{Attaching Stickers}\label{sec:attaching_stickers}

When training the generator, parameters will be updated according to the recognition results of human faces with created stickers.
Hence, the perturbed face images taken by cameras in FR systems should be effectively and accurately simulated.

Recent studies, which exploit adversarial stickers to deceive FR systems, proposed algorithms to digitally paste these stickers to human faces or facial accessories, which simulates the appearance of human faces with stickers in the real world~\cite{sharif2019general,pautov2019adversarial,komkov2019advhat}. However, these algorithms have apparent limitations:
they either assume that stickers are not bendable and thus cannot handle the situation
where the attacker does not face directly~\cite{sharif2019general},
or cannot digitally attach different shapes of stickers onto human faces~\cite{pautov2019adversarial,komkov2019advhat}.
The stickers crafted by FaceAdv have various shapes so that we have to design a new method to attach stickers to faces digitally.

\begin{figure}[t!]
    \centering
    \subfigure[The workflow of rendering the texture images]{
        \begin{minipage}[b]{0.45\textwidth}
        \centering
        \includegraphics[width=1.0\textwidth]{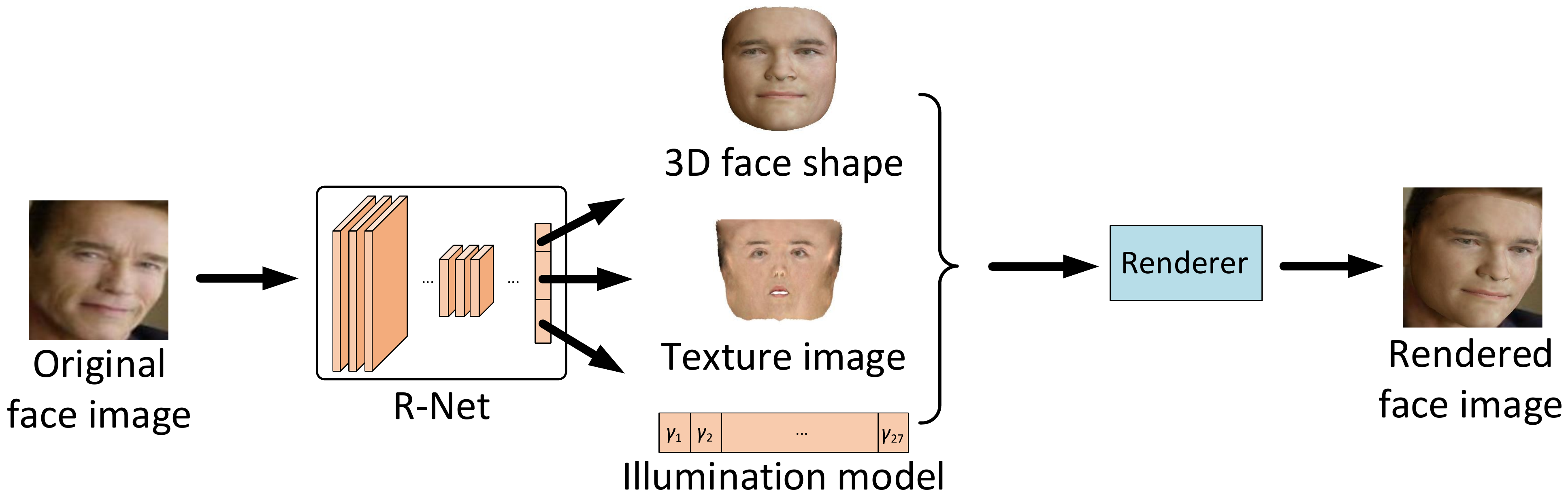}
        \label{fig:workflow}
        \end{minipage}
    }
    %%%%%

    \subfigure[The proposed RSO for digitally attaching stickers to faces]{
        \begin{minipage}{0.45\textwidth}
        \centering
        \includegraphics[width=1.0\textwidth]{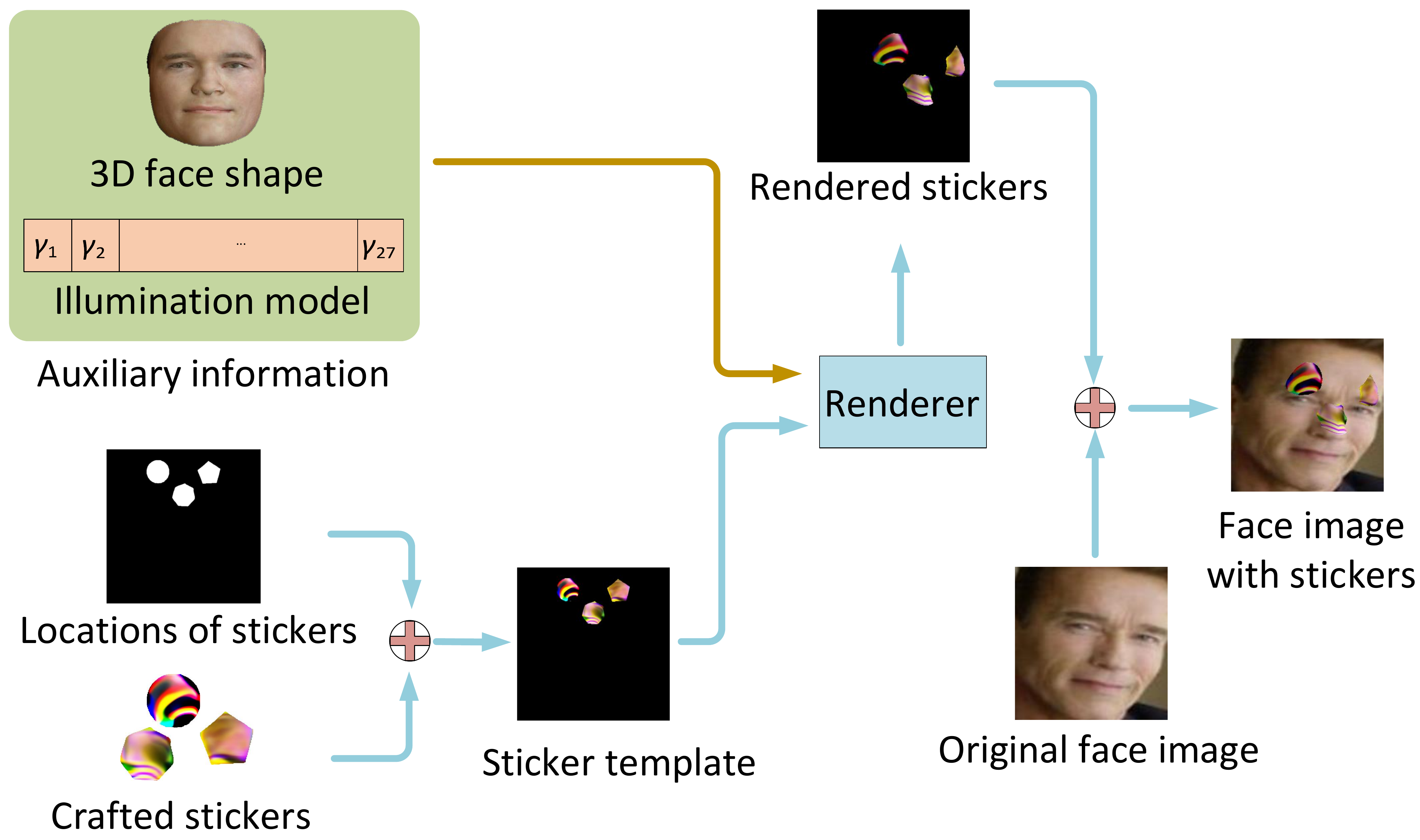}
        \label{fig:attaching}
        \end{minipage}
    }
    \caption{The implementation of the transformer $\mathcal{T}$. It utilizes R-Net to generate auxiliary information, and applies the worflow of rendering texture images to digitally paste adversarial stickers onto human faces.}
    \label{fig:transformer}
\end{figure}

We employ the 3D face reconstruction method R-Net~\cite{deng2019accurate} to estimate the 3D face shape, illumination model as well as the camera model,
as illustrated in Fig.~\ref{fig:workflow}, and leverage resulting face shapes to digitally attach adversarial stickers to human faces.
The detailed description of R-Net is deferred in Appendix~\ref{sec:3d_face_reconstruction}.
After getting these information, the differentiable renderer can render the texture image according to the 3D face shape while fusing the reckoned ambient brightness.

Motivated by the remarkable performance of R-Net, we present a method to digitally attach adversarial stickers onto human faces.
At the beginning, we directly edit the texture image (as shown in Fig.~\ref{fig:workflow}) by placing adversarial stickers on desired regions (i.e., two superciliary arches and the nasal bone), which is referred to as \emph{rendering whole image} (RWI).
However, from the rendered face image, we can find it loses many details (e.g., winkles) due to the limitations of the 3D face reconstruction. As a result, the recognition result cannot accurately measure the effectiveness of adversarial stickers.

To overcome this problem, we present a new method named  \emph{rendering stickers only} (RSO) to render the texture image with only stickers, as illustrated in Fig.~\ref{fig:attaching}.
This method replaces the texture image with the location image of stickers that indicates locations to place these stickers. The conversion from texture image to location image follows this rule: The pixel value in regions of the original texture image is set to $0$ while that in regions for pasting stickers is equal to $255$.
After putting these stickers on the chosen positions, the renderer can craft an image containing the stickers only, and the resulting image can cover the original face image to produce the face image with stickers.

In contrast with the rendered face image crafted by RWI,
the one created by RSO retains many realistic details and thereby is more similar to the original face image.
FaceAdv adopts RSO to digitally attach adversarial stickers on human faces in the transformer $\mathcal{T}$.
To reduce time consumption, FaceAdv generates the 3D face shape, the location image of stickers and the parameters of Spherical Harmonics using the R-Net in advance and stores them in the training set. By this way, the transformer $\mathcal{T}$ gains the auxiliary information from the input when training $\mathcal{G}$.

\subsection{Loss Functions}\label{sec:loss_designment}

Having the architecture of FaceAdv, in this subsection, we will design the loss functions for the two attack modes (i.e., dodging attacks and impersonating attacks) and formally describe the training algorithm.

Although GANs have achieved appealing results in image generation, training GANs stably is still a challenging problem. To alleviate this problem, we resort to Wasserstein GAN with gradient penalty (WGAN-GP)~\cite{gulrajani2017improved}, which is a variant of GANs that can be applied in GANs with different architectures.
Essentially, the goal of GANs is to transform the distribution of a random noise $n$ into the distribution of the input data (i.e., the shape image in this paper).
WGAN-GP utilizes $\mathcal{D}$ to calculate the Wasserstein distance between the shapes crafted by $\mathcal{G}$ and the real shapes, which is denoted by $\mathcal{L}_{GAN}$ in Fig.~\ref{fig:gans}.

The discriminator $\mathcal{D}$ aims to distinguish between the crafted shapes and the real shapes.
When feeding a shape into $\mathcal{D}$, it will output a number:
% {\color{red}if the number is below a threshold, the shape is judged as crafted by $\mathcal{G}$; otherwise, it is recognized as a real shape.}
the smaller number is more likely to be judged as crafted; otherwise, it is recognized as real.
Thus, the objective of $\mathcal{D}$ is minimizing $\mathcal{L}_{\mathcal{D}}$.
\begin{equation}\label{eq:objective_of_d}
    \begin{split}
        \mathcal{L}_{\mathcal{D}}(n, s) = &\mathbb{E}_{n \sim \mathbb{P}_n}(\mathcal{D}(\mathcal{G}(n))) \\
        &- \mathbb{E}_{s \sim \mathbb{P}_s}(\mathcal{D}(s)) \\
        &+ \lambda \mathbb{E}_{\hat{s} \sim \mathbb{P}_{\hat{s}}} \lbrack (\Vert \bigtriangledown_{\hat{s}} \mathcal{D}(\hat{s}) \Vert_2 - 1) ^2 \rbrack
    \end{split}
\end{equation}
where $\mathbb{P}_n$ and $\mathbb{P}_s$ are the distribution of the noise $n$ (i.e., the normal distribution) and the distribution of the real shape $s$, respectively.
The last term of Eq.~\eqref{eq:objective_of_d} is the gradient penalty, which makes training more stable.
$\mathbb{P}_{\hat{s}}$ is sampling uniformly along the straight lines between pairs of images sampled from the real shape distribution $\mathbb{P}_{s}$ and the crafted shape distribution $\mathbb{P}_{\mathcal{G}(n)}$, which can be formulated as:
\begin{equation}\label{eq:gradient_penalty}
    \hat{s} = \epsilon s + (1 - \epsilon) \mathcal{G}(n)
\end{equation}
where $s \sim \mathbb{P}(s)$, $\mathcal{G}(n) \sim \mathbb{P}_{\mathcal{G}(n)}$ and $\epsilon \sim U[0, 1]$.

The goal of $\mathcal{G}$ is to mislead $\mathcal{D}$ by labeling crafted shapes as real shapes, which can be expressed in Eq. \eqref{eq:gradient_penalty}.
\begin{equation}\label{eq:gradient_penalty}
    \mathcal{L}^s_\mathcal{G}(n) = - \mathbb{E}_{n \sim \mathbb{P}_n}(\mathcal{D}(\mathcal{G}(n)))
\end{equation}

% {\color{red}
% We refer to $\mathcal{L}_{\mathcal{D}}(n, s)$ and $\mathcal{L}^s_\mathcal{G}(n)$ as $\mathcal{L}_{GAN}$,
% because only the two losses can determine the performance of $\mathcal{G}$.}

$\mathcal{L}_{GAN}$ can be logically decomposed as $\mathcal{L}_{\mathcal{D}}(n, s)$ and $\mathcal{L}^s_\mathcal{G}(n)$, which can train $\mathcal{G}$ for crafting different shapes.

Next, we focus on the loss function $\mathcal{L}_{adv}$,
which represents either the opposite of the distance between predicted and ground-truth classes in dodging attacks,
or the distance between predicted and target classes in impersonating attacks.

In dodging attacks, the attacker $P_A$ aims to deceive the FR system $\mathcal{F}$ by misclassifying him as another person $P_B$ ($P_A \neq P_B$).
Thus, FaceAdv should reduce the probability of class $P_A$ and make it less than the probability of another class,
\begin{equation}\label{eq:dodging_attacks}
    \mathcal{L}_{adv}(x_A, n, a, P_A) = - \mathcal{F}(\mathcal{T}(x_A, \mathcal{G}(n), a), P_A)
\end{equation}
where $\mathcal{F}(\cdot, \cdot)$ represents the cross entropy that is commonly used in image classification~\cite{sharif2019general}.
By minimizing Eq.~\eqref{eq:dodging_attacks},
the output of this cross entropy function will increase so that the probability of the class $P_A$ can be reduced.

In impersonating attacks, we expect that FR recognizes the identity of the attacker $P_A$ as the target class $P_B$,
which can be formulated by Eq.~\eqref{eq:impersonating_attacks}.
\begin{equation}\label{eq:impersonating_attacks}
    \mathcal{L}_{adv}(x_A, n, a, P_B) = \mathcal{F}(\mathcal{T}(x_A, \mathcal{G}(n), a), P_B)
\end{equation}

In addition, there is another loss function considering the perturbation loss when transforming digital stickers to printed stickers.
These images captured by cameras in the real world comprise smooth and consistent patches, where colors change gradually~\cite{sharif2016accessorize}.
Due to the perturbation loss, extreme difference between adjacent pixels in adversarial stickers cannot be accurately printed by printers.
Consequently, adversarial stickers should be smooth.
We use the total variation loss to smooth these stickers,
which can be defined in Eq. \eqref{eq:total_variation},
\begin{equation}\label{eq:total_variation}
    \begin{split}
        \mathcal{L}_{tv}(n) = \sum_{i, j} \big( &(\mathcal{G}(n)_{i, j} - \mathcal{G}(n)_{i, j+1})^2 + \\
        &(\mathcal{G}(n)_{i, j} - \mathcal{G}(n)_{i+1, j})^2 \big)^{\frac{1}{2}}
    \end{split}
\end{equation}
where $\mathcal{G}(n)_{i, j}$ is the pixel in $\mathcal{G}(n)$ at coordinates $(i, j)$.
A lower value of $\mathcal{L}_{tv}(n)$ means the values of adjacent pixels are closer to each other and the printed stickers are more similar to the digital stickers.

Since $\mathcal{G}$ aims to craft stickers with different shapes to cheat the FR $\mathcal{F}$, the entire loss of $\mathcal{G}$ is defined in Eq. \eqref{eq:objective_of_g},
%by combing $\mathcal{L}_{\mathcal{G}}^s$, $\mathcal{L}_{adv}$ and $\mathcal{L}_{tv}$,
\begin{equation}\label{eq:objective_of_g}
    \mathcal{L}_{\mathcal{G}} = \mathcal{L}_{\mathcal{G}}^s + \alpha \mathcal{L}_{adv} + \beta \mathcal{L}_{tv}
\end{equation}
where $\alpha$ and $\beta$ are weights that control the relative importance of each objective.
When $\alpha$ is larger, the shape of created stickers will be uncontrollable and is difficult to be cut out in the real world, while a larger $\beta$ means the color in the generated stickers trends to be the same.
% {\color{red}Thus, it is important to select a reasonable value for $\alpha$ and $\beta$.}
Thus, we conduct experiments to select a reasonable value for $\alpha$ and $\beta$.

The algorithm for training FaceAdv is depicted in Appendix~\ref{sec:training_generator}.
The output of training FaceAdv is an sticker generator to craft a number of stickers with different shapes that can fool the target FR system $\mathcal{F}$.

\section{Performance Evaluation}\label{sec:evaluation}
%----------------------------------------------------- Evaluation

In this section, we comprehensively evaluate the performance of the proposed FaceAdv against 3 stat-of-the-art FR systems.
We are dedicated to answering the following questions: 1) How does FaceAdv generate appropriate stickers for each target FR system? 2) Will FaceAdv outperform other methods when launching dodging attacks and impersonating attacks?
3) How will potential influencing factors affect the performance of FaceAdv?
%First, we describe the default settings that are used for evaluation. Then, we introduce the experimental protocol. Lastly, the experimental results are presented and discussed. %All experimental codes will be released upon request.

\subsection{Experimental Settings}\label{sec:experimental_settings}

\subsubsection{Testbed}

We select ArcFace~\cite{deng2019arcface}, CosFace~\cite{wang2018cosface} and FaceNet~\cite{schroff2015facenet} as the state-of-the-art feature extractors in FR systems.
Given a specific feature extractor, the corresponding MLP classifier of the target FR system needs to be trained.
We use a server with 32GB RAM, Nvidia RTX 2070 GPU and AMD Ryzen 7 7200X CPU for all training tasks.
The target FR systems are deployed on a PC with 16GB RAM and Intel Core i7-9750H CPU.
The camera is the Logitech C270 and captures images with 960$\times$1280 in pixels.
The printer for making physical stickers is HP DeskJet 2677.
In addition, the light source is Philips 66135, which can change the ambient light to 30lux, 130lux and 250lux.
The diagrammatic sketch of the experimental site is shown in Appendix~\ref{sec:experimental_site}.

The default values of the experimental parameters are set as follows: user-camera distance of $50$cm, sticker size of $90 \times 90$ in pixels, environmental brightness of $130$lux,
$r$ of $5e^{-4}$, $\alpha = 100$, $\beta = 1$ and a head pose of facing directly forward. All the experiments are conducted using the default settings unless otherwise specified.

\subsubsection{Baselines}

We employ AGNs~\cite{sharif2019general} as the state-of-the-art physical-world attack for comparison.
It is because both FaceAdv and AGNs use sticker-like perturbations and adopt the similar methodology (i.e., GANs) in generating perturbations.
In AGNs, a parameter $\kappa$ is used to balance effectiveness and inconspicuousness, e.g., when $\kappa$ is close to zero, the crafted stickers pay more attention on effectiveness.
As FaceAdv gives priority to effectiveness, to make a fair comparison, we set $\kappa$ as $0.25$ to relax the restriction on the stealthiness of stickers.

\subsubsection{Datasets}

The well-known face dataset LFW~\cite{LFWTech} is a public benchmark for FR systems, which consists of $13,233$ images from $5,749$ individuals.
For ArcFace~\cite{arcface_model}, CosFace~\cite{cosface_model} and FaceNet~\cite{facenet_model}, their pre-trained models based on PyTorch are publicly available, which achieve an accuracy of $99.65\%$, $99.23\%$, $99.65\%$ on LFW, respectively.
As mentioned above, we need to train a MLP classifier, which follows a feature extractor to form a complete FR system.
However, we find that $4,069$ labels in LFW only have a single image and thus cannot be used to train the MLP classifier. Therefore, we only select the labels with more than $10$ images, which results in a subset $\mathbf{LFW^-}$ containing $143$ labels as the \emph{victim} dataset.

\begin{table}[t]
    \caption{Details of datasets}
    \label{tab:train_dataset}
    \centering
    \scalebox{0.8}{
        \renewcommand\arraystretch{1.3}
        \begin{tabular}{|c|c|c|c|c|c|} \hline
        Dataset & \# Labels & \# Images & ArcFace & CosFace & FaceNet \\ \hline
        $\rm{LFW}$ & $5,749$ & $13,233$ &  $99.65\%$ & $99.23\%$ & $99.65\%$ \\ \hline
        $\rm{LFW^-}$ & $143$ & $4,174$ &  $97.51\%$ & $97.28\%$ & $99.57\%$ \\ \hline
        VulFace & $10$ & $450$ & $99.80\%$ & $99.41\%$ & $100.00\%$ \\ \hline
        \end{tabular}
    }
\end{table}

For launching physical-world attacks, $10$ student volunteers ($6$ males and $4$ females, aged between 20 and 25) are involved in collecting the \emph{attacker} dataset \textbf{VulFace}, which contains $45$ face images for each volunteer.
Then, we apply $80\%$ of $\rm{LFW^-}$ and VulFace to train the MLP classifier, and the rest for testing the target FR systems. The accuracy of the FR systems with ArcFace, CosFace, FaceNet is shown in Table~\ref{tab:train_dataset}.

\begin{figure}
    \centering
    \includegraphics[width=0.7\linewidth]{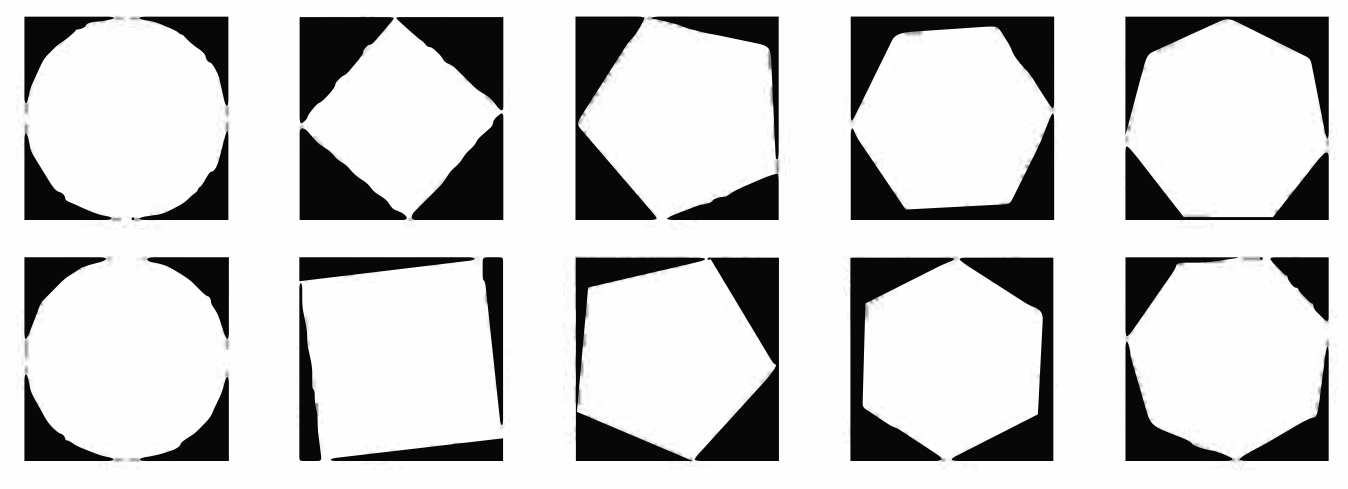}
    \caption{Sampled shapes in the FiveShape.}
    \label{fig:fiveshape}
\end{figure}

We also have a \emph{shape} dataset \textbf{FiveShape} to train the generator $\mathcal{G}$ to fabricate different shapes, which contains $5$ different shapes (i.e., circle, square, pentagon, hexagon, heptagon).
This dataset consists of $15,000$ images and some samples in FiveShape are illustrated in Fig.~\ref{fig:fiveshape}.
There is the trick that there should not is a big difference between the area of different shapes in the shape dataset. Assuming the pentagram is in the shape dataset, since cutting the square into a pentagram loses a large area of the adversarial sticker that will damage the attack success rate, the generator $\mathcal{G}$ will not generate stickers with the pentagram shape.

%Generally, it takes a long time to train GANs and the goal of GANs in FaceAdv is not to create stickers to cheat FR systems but to craft different shape masks.
% In order to reduce the training time of $\mathcal{G}$, we apply FiveShape to pretrain $\mathcal{G}$ and $\mathcal{D}$ only through $\mathcal{L}_{GAN}$ so that the shape part of the $\mathcal{G}$ can generate different shapes before training the sticker part of $\mathcal{G}$.

\subsubsection{Performance Metrics}

In the real-world, it is hard for an attacker to keep still when launching certain attacks, leading to different attacking results.
To address this problem, for each attack, we record a video of $20$ seconds and extract $110$ frames on average to obtain the classification result of the target FR systems.
For dodging attacks, the success rate is the fraction of frames that are classified as a different person from the attacker, and for impersonating attacks, it is defined by the fraction of frames that are classified as the target person.

\subsection{Evaluation of Sticker Settings}

The settings of adversarial stickers (e.g., their locations and size) are critical to determine the performance of FaceAdv.
Since launching physical-world attacks are more time-consuming, we resort to digital attacks in the following experiments:
the adversarial images are created by attaching stickers on face images of the attackers in VulFace (i.e., 45 images for each attacker) and then used to cheat the target FR systems. Only the impersonate attack mode is involved as it is more challenging than the dodging attack mode.

\subsubsection{Locations of Stickers}\label{sec:locations_of_stickers}

As mentioned in Section~\ref{sec:sticker_localization}, we use Guided Grad-CAM to select $5$ critical regions for attaching adversarial stickers.
Since FaceAdv can generate $3$ different stickers at a time, there are 10 combinations\footnote{We name these combinations from \#$1$ to \#$10$, which consists of taking $3$ from $5$ candidate positions (i.e., right superciliary arch, left superciliary arch, nasal bone, right nasolabial sulcus and left nasolabial sulcus) from left to right in turn.} of candidate positions to paste stickers.
We randomly select $5$ attackers from VulFace and $2$ victims from $\rm{LFW^-}$, and test each combination in turn for each attacker-victim pair.
Thus, there are altogether $5 \times 2 \times 45 \times 10= 4500$ tests for each target FR system.
In these experiments, the sticker size is fixed to $90\rm{px} \times 90\rm{px}$.

\begin{figure}[t!]
    \centering
    \includegraphics[width=0.7\linewidth]{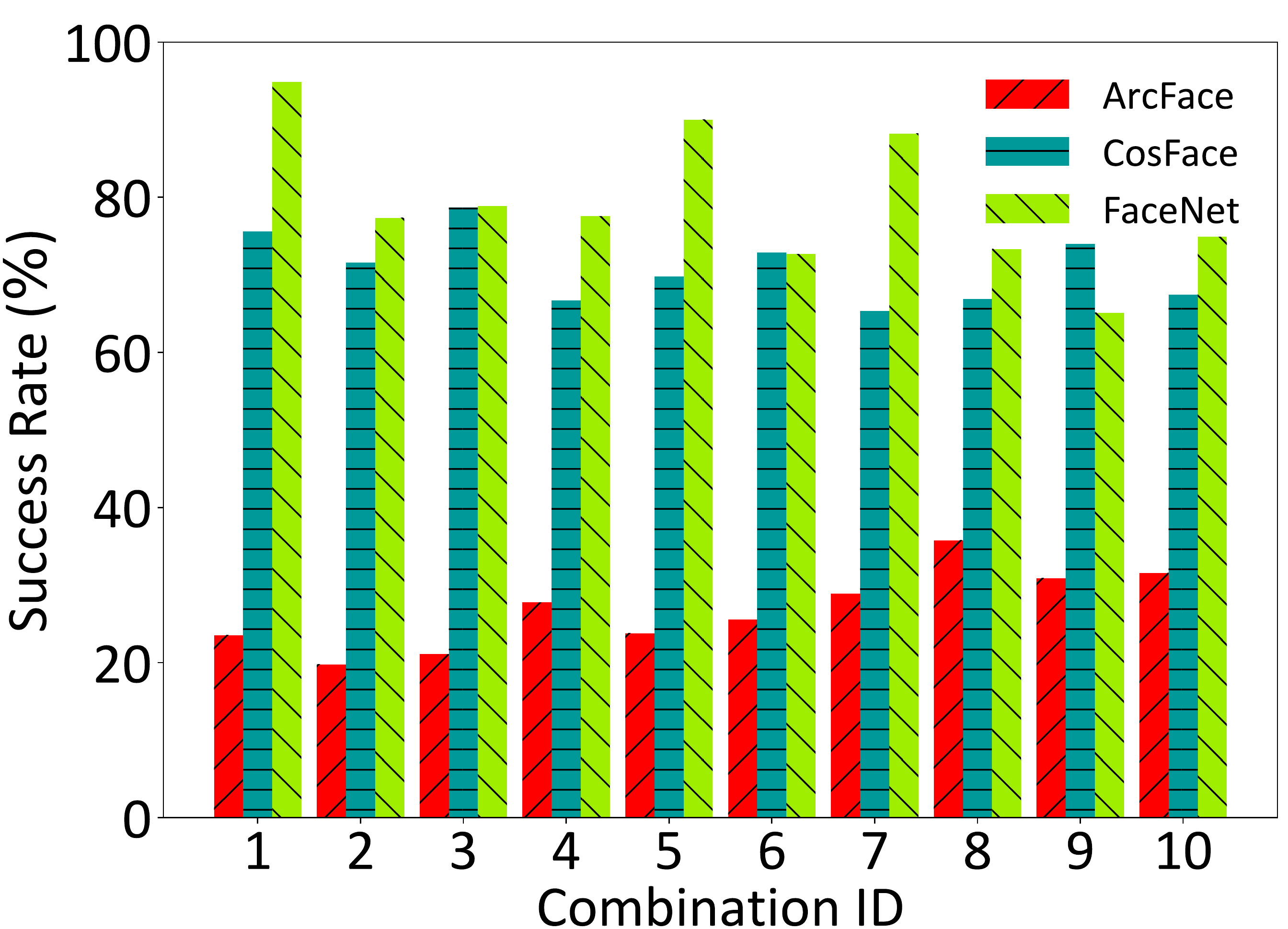}
    \caption{Success rates of different location combinations.}
    \label{fig:success_rate_different_locations}
\end{figure}

The success rates of FaceAdv are summarized in Fig.~\ref{fig:success_rate_different_locations}.
In general, FaceNet is more vulnerable than the other two FR systems.
We also observe that the combination that achieves the highest success rates varies a lot among different FR systems.
Based on the results, we opt combination \#$8$ (i.e., left superciliary arch, nasal bone, left nasolabial sulcus), \#$3$ (i.e., right superciliary arch, left superciliary arch, left nasolabial sulcus), and \#$1$ (i.e., right superciliary arch, left superciliary arch. nasal bone) as the best choices for attacking ArcFace, CosFace and FaceNet, respectively.

This result can be explained by the localization maps in Table.~\ref{tab:grad_cam}.
FaceNet always pays more attention to five sense organs (e.g., eyes, the nose, the mouth) and ignores the nasolabial sulcus, which is exactly the positions of combination \#1.
In contrast, both ArcFace and CosFace obtain information from the nasolabial sulcus.
In addition, ArcFace tends to extract face features from partial face region, which reduces the effectiveness of stickers.
In the rest of the experiments, we regard these position combinations as the default settings, unless otherwise specified.

\subsubsection{Size of Stickers}\label{sec:size_of_stickers}

We investigate how the size of stickers affects the performance of FaceAdv.
% {\color{red}
% Unlike other methods using stickers to deceive the FR systems, in FaceAdv, stickers will be placed on the sticker template and then rendered onto face images of attackers, as mentioned in Fig.~\ref{fig:attaching}.
% Since the size of stickers in the final face images will vary with the head pose, in this paper, sticker size refers to the size in the sticker template, rather than in the face images with stickers.}
Since the face size of each person is not necessarily the same and faces of different persons will be scaled to the same resolution in the sticker template, we measure the sticker size by pixel (i.e., px). The size of stickers in final face images will vary with the head pose, as illustrated in Fig.~\ref{fig:attaching}, and thus sticker size refers to the size in the sticker template rather than in the face images with stickers.

\begin{table}[t]
    \caption{Success rates with varying sticker sizes}
    \label{tab:different_sticker_size}
    \centering
    \renewcommand\arraystretch{1.2}
    \scalebox{0.8}{
        \begin{tabular}{|c|c|c|c|c|c|c|c|c|c|c|} \hline
        \multirow{2}{*}{Target Model} & \multicolumn{3}{c|}{Sticker Size (px)} \\
        \cline{2-4}
        ~ & $80 \times 80$ & $90 \times 90$ & $100 \times 100$ \\ \hline
        ArcFace & $16.89\%$ & $35.78\%$ & $56.67\%$ \\ \hline
        CosFace & $58.00\%$ & $78.67\%$ & $89.33\%$  \\ \hline
        FaceNet & $89.78\%$ & $94.89\%$ & $99.78\%$  \\ \hline
        \end{tabular}
    }
\end{table}

% In the sticker template, faces of different persons will be scaled to the same resolution. Since the face size of each person is not necessarily the same, we measure the sticker size by pixel (i.e., $\rm{px}$).

We use the same attacker-victim pairs as in Section \ref{sec:locations_of_stickers} for evaluation.
We consider $3$ sizes: $80\rm{px} \times 80\rm{px}$, $90\rm{px} \times 90\rm{px}$ and $100\rm{px} \times 100\rm{px}$.
For each target FR system, we conduct $5 \times 2 \times 45 \times 3 = 1350$ tests.

In our implementation, the size of original face images, sticker templates, and perturbed face images are respectively $300\rm{px} \times 300\rm{px}$, $600\rm{px} \times 600\rm{px}$ and $300\rm{px} \times 300\rm{px}$.
Because ArcFace, CosFace and FaceNet separately take images with $112\rm{px} \times 112\rm{px}$, $96\rm{px} \times 112\rm{px}$ and $160\rm{px} \times 160\rm{px}$ resolution as input, the perturbed images will be scaled into a certain resolution.
Therefore, the total area of adversarial stickers accounts for less than $10\%$ of that of sticker templates.

The results are shown in Table~\ref{tab:different_sticker_size}.
In general, a larger sticker help improve the success rate.
In the physical world, however, the length of nasal bone and the distance between facial features limit the size of stickers.
If the size of stickers pasted on the nasal bone is too large, it will cover eyes and cannot commendably handle the arc between the nasal bone and the glabella.
Due to this limitation, we select $90\rm{px} \times 90\rm{px}$ as the default sticker size.

\subsubsection{Time Efficiency}

In order to reduce the training time of $\mathcal{G}$, we apply FiveShape to pretrain $\mathcal{G}$ and $\mathcal{D}$ only through $\mathcal{L}_{GAN}$ so that the shape part of $\mathcal{G}$ can generate different shapes before training the sticker part of $\mathcal{G}$.

With the fixed location and size of stickers,
FaceAdv can train the generator $\mathcal{G}$ only once to generate adversarial stickers for a specific FR system.
In our tests, FaceAdv takes less than $26$ minutes on average to train $\mathcal{G}$ and less than 2 seconds to craft adversarial stickers. When upgrading GPU to Nvidia RTX 2080Ti, the corresponding time is reduced to less than $20$ minutes and less than $1$ second, respectively. After stickers are crafted by $\mathcal{G}$, these stickers can be printed, tailored, and pasted on the selected facial regions of an attacker.

\subsection{Evaluation of Dodging and Impersonating Attacks}\label{sec:dodging_and_impersonating_attacks}

In Table~\ref{tab:case_study}, we present several adversarial examples for dodging and impersonating attacks against the target FR systems.
In this subsection, we will conduct extensive experiments to evaluate the effectiveness of FaceAdv.

The method AGNs utilizes $7$ green marks to indicate the location of eyeglasses, and it cannot work well when an attacker does not look straight ahead, because the green marks may be cut off by the face detector.
In order to make a fair comparison, we take another $20$ face images for each volunteer who looks straight ahead, and these images are only used to calculate the success rate in the digital world. In the physical world, the user-camera distance is $50cm$, the environmental luminance is $130$lux, and the head pose is straight ahead.

AGNs was evaluated in its original paper~\cite{sharif2019general} with a FR system named VGGFace~\cite{parkhi2015deep}.
Thus, we conduct preliminary experiments to demonstrate that our reproduction of AGNs can achieve similar results reported
in~\cite{sharif2019general}.
The results are shown in Appendix~\ref{sec:attacking_vggface}, which shows that both FaceAdv and AGNs can successfully attack VGGFace in the digital and physical world.
Since VGGFace has been proved to be inferior to the 3 FR systems (i.e., ArcFace, CosFace and FaceNet) in the large-scale dataset~\cite{guo2016ms}, we will leverage these systems for evaluation in the following experiments.

\subsubsection{Dodging Attacks}

We evaluate success rates of dodging attacks with two methods in the digital and physical world, as shown in Table~\ref{tab:attack_results}.
%When conducting this experiment, all volunteers in VulFace are chosen as attackers, which results in a total of $10 \times 3 \times 2 = 60$ tests for each target FR system.

In digital scenarios, we use FaceAdv or AGNs to generate stickers and digitally attach them onto the 20 face images of each attacker, which results in a total of $10 \times 20 = 200$ tests for each target FR system.
In physical scenarios, each attacker has 110 images captured from a 20-second video, and we have $10 \times 110 = 1100$ tests for each target FR system.

\begin{table}[t]
    \caption{Success rate in dodging or impersonating attacks}
    \label{tab:attack_results}
    \centering
    \renewcommand\arraystretch{1.3}
    \scalebox{0.76}{
        \begin{tabular}{|c|c|c|c|c|c|} \hline

        \multirow{2}{*}{Attack Mode} & \multirow{2}{*}{Scenario} & \multirow{2}{*}{Method} & \multicolumn{3}{c|}{Target FR System} \\
        \cline{4-6}
        ~ & ~ & ~ & ArcFace & CosFace & FaceNet \\ \hline
        \multirow{4}{*}{Dodging} & \multirow{2}{*}{Digital} & FaceAdv & $84.00\%$ & $100.00\%$ & $100.00\%$ \\
        \cline{3-6}
        ~ & ~ & AGNs & $86.50\%$ & $99.50\%$ & $93.00\%$ \\
        \cline{2-6}
        ~ & \multirow{2}{*}{Physical} & FaceAdv & $75.00\%$ & $100.00\%$ & $100.00\%$  \\
        \cline{3-6}
        ~ & ~ & AGNs & $50.00\%$ & $50.00\%$ & $62.50\%$  \\ \hline
        \multirow{4}{*}{Impersonating} & \multirow{2}{*}{Digital} & FaceAdv & $63.17\%$ & $88.67\%$ & $94.50\%$ \\
        \cline{3-6}
        ~ & ~ & AGNs & $14.83\%$ & $70.83\%$ & $25.67\%$  \\
        \cline{2-6}
        ~ & \multirow{2}{*}{Physical} & FaceAdv &$4.17\%$ & $29.17\%$ & $54.17\%$ \\
        \cline{3-6}
        ~ & ~ & AGNs & $0.00\%$ & $0.00\%$ & $0.00\%$  \\ \hline
        \end{tabular}
    }
\end{table}

We have \emph{two key observations} from the attack results.

\emph{First}, the success rate in the digital world is higher than that in the physical world, as shown in Table~\ref{tab:attack_results}. When attacking ArcFace in the physical world, the success rate of FaceAdv drops about $9\%$. As for AGNs, the success rate is reduced by $30\%$ on average. This confirms that the perturbation loss reduces the effectiveness of adversarial examples.

\emph{Second}, FaceAdv achieves higher success rates in both digital and physical scenarios.
Compared with FaceAdv, AGNs has severe perturbation loss in physical scenarios, which results in significate performance degradation ($\sim30\%$) in the physical world.
This demonstrates that the sticker generator and the transformer in FaceAdv successfully simulate perturbed images with stickers pasted on real human faces.

\subsubsection{Impersonating Attacks}

For launching impersonating attacks, we randomly select 3 victims from $\mathbf{LFW^-}$ and employ all volunteers in VulFace as attackers.
As a result, for each target FR system, there are altogether $10 \times 20 \times 3 = 600$ tests in digital scenarios and $10 \times 110 \times 3 = 3300$ tests in physical scenarios.

The attack success rates are summarized in Table~\ref{tab:attack_results}.
There are \emph{three key observations} from the results.

\emph{First}, the success rates of FaceAdv are higher than those of AGNs by a large margin in both digital and physical scenarios. It demonstrates the proposed method is superior to AGNs.

\emph{Second}, compared with the digital world, the success rate of FaceAdv in the physical world reduces by $50\%$ on average.
Launching impersonating attacks is more difficult than performing dodging attacks, as the recognition result is the pre-determined victim.

\emph{Third}, AGNs can successfully attack FR systems in the digital world, but fail in the physical world.
The reason is two-fold:
1) the area of eyeglass frames is too small to attack FR systems, as the size of stickers can greatly affect the performance (Section~\ref{sec:size_of_stickers});
2) the feature extractor is trained on the large-scale dataset (e.g., MS-Celeb-1M~\cite{guo2016ms}) so that it can resist potential adverarial attacks.

A surprising observation is that attackers, when facing directly the camera, can hardly attack ArcFace.
This reveals that a normal head pose may not achieve the highest success rate, which motivates us to conduct further investigation on the influence of head poses in Section~\ref{sec:head_pose}.

\subsection{Evaluation on Influencing Factors}

To investigate the performance of FaceAdv in different conditions, we investigate the success rate of FaceAdv by varying several influencing factors, including the user-camera distance, the luminance, and the head pose.
In the following experiments, we also use the $10$ volunteers as attackers and the $3$ individuals as victims.

\subsubsection{User-Camera Distance}

\begin{table}[t]
    \caption{Success rate with varying distance}
    \label{tab:attack_results_varying_distance}
    \centering
    \renewcommand\arraystretch{1.3}
    \scalebox{0.8}{
        \begin{tabular}{|c|c|c|c|c|} \hline

        \multirow{2}{*}{Mode} & \multirow{2}{*}{Target Model} & \multicolumn{3}{c|}{User-Camera Distance (cm)} \\
        \cline{3-5}
        ~ & ~ & 30 & 50 & 70 \\ \hline
        \multirow{3}{*}{Dodging} & ArcFace & $87.50\%$ & $75.00\%$ & $37.50\%$ \\
        \cline{2-5}
        ~ & CosFace & $100.00\%$ & $100.00\%$ & $100.00\%$ \\
        \cline{2-5}
        ~ & FaceNet & $100.00\%$ & $100.00\%$ & $100.00\%$ \\ \hline
        \multirow{3}{*}{Impersonating} & ArcFace & $4.17\%$ & $4.17\%$ & $8.33\%$ \\
        \cline{2-5}
        ~ & CosFace & $29.17\%$ & $29.17\%$ & $29.17\%$ \\
        \cline{2-5}
        ~ & FaceNet & $45.83\%$ & $54.17\%$ & $33.33\%$ \\ \hline
        \end{tabular}
    }
\end{table}

In physical scenarios, attackers (i.e., users) cannot precisely control the distance to the camera, which requires crafted stickers can work with different distances.
Thus, we evaluate the success rates of FaceAdv with user-camera distance of 30cm (e.g., unlocking mobile phone or laptop), 50cm (e.g., passing the access control of buildings) and 70cm (e.g., using face scan payment). The $50$cm is the default setting.
The results are shown in Table~\ref{tab:attack_results_varying_distance}.

From these results, we can find that adversarial stickers crafted by FaceAdv work stably in both digital and physical scenarios.
This is mainly because each time when training $\mathcal{G}$, FaceAdv attaches adversarial stickers to multiple face images captured at different distances, as described in Section~\ref{sec:sticker_generator}.

In the physical world, the size of face images is different with varying distances so that the FR system will rescale face images to the resolution of input images (i.e., $112\rm{px} \times 112\rm{px}$ in ArcFace, $96\rm{px} \times 112\rm{px}$ in CosFace, $160\rm{px} \times 160\rm{px}$ in FaceNet). FaceAdv simulates this process when training $\mathcal{G}$. The size of final face images with stickers is $300\rm{px} \times 300\rm{px}$, and FaceAdv resamples them to the certain resolution, which ensures stickers are robust with the image resampling.

\subsubsection{Brightness Level}

\begin{table}[t]
    \caption{Success rate with varying luminance}
    \label{tab:attack_results_varying_luminance}
    \centering
    \renewcommand\arraystretch{1.3}
    \scalebox{0.8}{
        \begin{tabular}{|c|c|c|c|c|} \hline

        \multirow{2}{*}{Mode} & \multirow{2}{*}{Target Model} & \multicolumn{3}{c|}{The Level of Brightness (lux)} \\
        \cline{3-5}
        ~ & ~ & 30 & 130 & 250 \\ \hline
        \multirow{3}{*}{Dodging} & ArcFace & $50.00\%$ & $75.00\%$ & $75.00\%$ \\
        \cline{2-5}
        ~ & CosFace & $100.00\%$ & $100.00\%$ & $100.00\%$ \\
        \cline{2-5}
        ~ & FaceNet & $100.00\%$ & $100.00\%$ & $100.00\%$ \\ \hline
        \multirow{3}{*}{Impersonating} & ArcFace & $4.17\%$ & $4.17\%$ & $0.00\%$ \\
        \cline{2-5}
        ~ & CosFace & $29.17\%$ & $29.17\%$ & $33.33\%$ \\
        \cline{2-5}
        ~ & FaceNet & $29.17\%$ & $54.17\%$ & $41.71\%$ \\ \hline
        \end{tabular}
    }
\end{table}

The luminance can also influence the performance of adversarial stickers. We evaluate the success rates of FaceAdv with varying environment brightness, i.e., 30lux, 130lux (by default) and 250lux. They represent weak indoor light, sunny day, and strong indoor light, respectively. The results are shown in Table~\ref{tab:attack_results_varying_luminance}.

In general, the performance of FaceAdv changes slightly with varying environment brightness.
This is because we employ R-Net to estimate the environment luminance in Section~\ref{sec:attaching_stickers}. When digitally attaching crafted stickers onto face images, FaceAdv will change the brightness of these stickers to fit that of face images according to the parameters of illumination model acquired in advance.
The aim of this process is to simulate the appearance of stickers in different luminance so as to improve the robustness of adversarial stickers.

However, we can find FaceAdv fail when attacking ArcFace in the environment brightness of 250lux. On the one hand, the success rate of attacking ArcFace is relatively low in other conditions, which indicates the difficulty in attacking this FR system.
On the other hand, it also shows that the illumination model still needs to be improved.

\subsubsection{Head Pose}\label{sec:head_pose}

\begin{table}[t]
    \caption{Success rate with varying the head pose}
    \label{tab:attack_results_varying_head_pose}
    \centering
    \renewcommand\arraystretch{1.3}
    \scalebox{0.62}{
        \begin{tabular}{|c|c|c|c|c|c|c|} \hline

        \multirow{2}{*}{Mode} & \multirow{2}{*}{Target Model} & \multicolumn{5}{c|}{Head Pose} \\
        \cline{3-7}
        ~ & ~ & HN & HL & HR & HU & HB \\ \hline
        \multirow{3}{*}{Dodging} & ArcFace & $75.00\%$ & $75.00\%$ & $87.50\%$ & $76.25\%$ & $87.50\%$  \\
        \cline{2-7}
        ~ & CosFace & $100.00\%$ & $100.00\%$ & $100.00\%$ & $100.00\%$ & $100.00\%$ \\
        \cline{2-7}
        ~ & FaceNet & $100.00\%$ & $87.50\%$ & $100.00\%$ & $100.00\%$ & $100.00\%$ \\ \hline
        \multirow{3}{*}{Impersonating} & ArcFace & $4.17\%$ & $0.00\%$ & $29.17\%$ & $12.50\%$ & $8.33\%$ \\
        \cline{2-7}
        ~ & CosFace & $29.17\%$ & $16.70\%$ & $33.33\%$ & $41.67\%$ & $25.00\%$ \\
        \cline{2-7}
        ~ & FaceNet & $54.17\%$ & $16.67\%$ & $29.17\%$ & $16.67\%$ & $45.83\%$ \\ \hline
        \end{tabular}
    }
\end{table}

As discussed earlier, FaceAdv achieves the lower success rate of attacking ArcFace when volunteers faces the camera directly.
We conduct a series of experiments to investigate the success rate of FaceAdv with varying head poses.

We select typical head poses: normal (HN, the default head pose), turning head to the left by 20deg (HR) or to the right by 20deg (HL), raising up head by 20deg (HU) or lowering head by 20deg (HB). The success rates with these head poses are shown in Table~\ref{tab:attack_results_varying_head_pose}.

FaceAdv achieves the highest success rate ($29.17\%$)
on ArcFace when the head pose is HR.
In Section~\ref{sec:locations_of_stickers}, we select the combination \#8 (i.e., left superciliary arch, nasal bone left, nasolabial sulcus) to attach stickers, which achieves the best average performance under different conditions.
However, locations of the combination \#8 are mainly on the left side of the face. Thus, when the head pose of volunteers turns to HR, the stickers attached on the left side are totally exposed to the camera, and the success rate becomes higher than that in other directions. Apparently, the stickers are basically invisible to the camera in HL, resulting the worst performance.

The performance of FaceAdv is relatively stable for attacking the other two FR systems. This is because we adopt two measures to improve the robustness of FaceAdv with varying head poses. First, we propose the new method based on 3D face reconstruction to digitally attach stickers onto face images.
Second, each time when training $\mathcal{G}$, FaceAdv attaches these stickers onto images captured from different head poses and then feeds them into the target FR system.

\section{Discussion}\label{sec:discussion}
%----------------------------------------------------- Discussion

The results of our study show that FaceAdv is an effective algorithm for generating physical-world adversarial examples against the state-of-the-art FR systems.
% Based on our observations during evaluations,
Next, we discuss several directions that can improve the performance of FaceAdv.

\textbf{Sticker Size}.
The size and location of printed stickers will effect the performance. Ideally, stickers should be the same as those in the digital world. However, face images are captured in different directions and the size of human faces is also different, making the size of printed stickers hardly estimated. We refer to this gap as \emph{fabrication error}.

We randomly resample stickers before feeding them into the transformer $\mathcal{T}$ to simulate the difference of sticker size and location between digital and physical scenarios. Specifically, we stochastically scale ($0.9 \sim 1.1$), rotate ($-10\deg \sim 10\deg$) and translate ($-10\rm{px} \sim \rm{px}$) stickers to imitate the error, which improves the robustness of FaceAdv.

\textbf{Transformer}. The performance of the transformer $\mathcal{T}$ determines the difference of face images with stickers between digital scenarios and physical scenarios.
A more effective algorithm for 3D face reconstruction will improve the performance of FaceAdv.

An alternative choice is to abandon $\mathcal{T}$ and directly place stickers on face images.
In that case, the renderer calculates the sticker template in reverse, and cuts physical stickers out. However, the shape of physical stickers is irregular because of the radian of face, which enhances the difficulty of tailoring stickers.
In addition, if $\mathcal{T}$ is removed, the fabrication error will cause significant performance degradation.
Therefore, we use the methodology described in Fig. \ref{fig:gans} rather than this alternative choice.

\textbf{Meaningful contents}.
% We introduce GANs to constrain the shape of stickers. However,
GANs can produce real-looking stickers when guaranteeing attack success rate~\cite{sharif2019general}. The discriminator is used to distinguish between fabricated stickers and real eyeglass-frame designs, which means the real pattern can also attack the FR system.

We try some datasets (e.g., cartoon character datasets, national flag datasets) to train the generator so that crafted stickers have certain physical meanings to reduce the attention of people nearby. However, all these attempts fail, because of the trade-off between fidelity and effectiveness.
In addition, the results in Section~\ref{sec:dodging_and_impersonating_attacks} also show that constraining the content of stickers will influence the effectiveness.

\section{Conclusion}\label{sec:conclusion}
%----------------------------------------------------- Conclusion

In this paper, we proposed a method named FaceAdv to automatically generate adversarial stickers, misleading the results of FR systems in the physical world. We employed an architecture of GANs to train a generator to craft adversarial stickers so as to fabricate a large number of stickers with different shapes after training, and proposed a novel method to digitally attach these stickers onto face images. Extensive experimental results demonstrated that FaceAdv can achieve the high success rate in physical scenarios with different environmental conditions. In future work, we will further investigate techniques to improve the effectiveness of FaceAdv and explore powerful defenses.

% \newpage
{\small
\bibliographystyle{abbrv}
\bibliography{refs}
}
\appendix

\section{The Architecture of GANs}\label{sec:architecture_of_gans}

FaceAdv utilizes the GANs to craft adversarial stickers, which can be divided into two parts: the generator $\mathcal{G}$ and the discriminator $\mathcal{D}$. $\mathcal{G}$ fabricates three pairs of stickers and shapes, and $\mathcal{D}$ is utilized to distinguish whether the input shape image is from the real dataset or not for guaranteeing that $\mathcal{G}$ can generate shape images with the same as that of shape dataset.

$\mathcal{G}$ can be splited into 3 identical branches for generating 3 adversarial stickers with different shapes. Each branch is composed of two parts for generating stickers and shape masks. However, in order to reduce the number of parameters, the first half layers of each part is the same, which are called \textit{Basic Layers}, and the second half layers are separately called \textit{Shape Layers} and \textit{Sticker Layers}.

\begin{table}[t]
    \caption{The architecture of $\mathcal{G}$}
    \label{tab:architecture_of_the_generator}
    \centering
    \renewcommand\arraystretch{1.3}
    \scalebox{0.65}{
      \begin{tabular}{|p{3cm}<{\centering}|p{4cm}<{\centering}|p{4cm}<{\centering}|} \hline
        ~ & Sticker Branch & Shape Branch \\ \hline
        Input Shape & \multicolumn{2}{c|}{$1 \times 32$} \\ \hline
        \multirow{9}{*}{Basic Layers} & \multicolumn{2}{c|}{FC($32 \times 16000$)} \\
        ~ & \multicolumn{2}{c|}{BatchNorm1d($16000$)} \\
        ~ & \multicolumn{2}{c|}{ReLU} \\
        ~ & \multicolumn{2}{c|}{Reshape($5 \times 5 \times 640$)} \\
        \cline{2-3}
        ~ & \multicolumn{2}{c|}{BatchNorm2d} \\
        ~ & \multicolumn{2}{c|}{ReLU} \\
        ~ & \multicolumn{2}{c|}{UpSample} \\
        ~ & \multicolumn{2}{c|}{(repeat twice)} \\
        \cline{2-3}
        ~ & \multicolumn{2}{c|}{Conv2d($20 \times 20 \times 160$)} \\ \hline
        \multirow{7}{*}{Sticker/Shape Layers} & \multicolumn{2}{c|}{BatchNorm2d} \\
        ~ & \multicolumn{2}{c|}{ReLU} \\
        ~ & \multicolumn{2}{c|}{UpSample} \\
        ~ & \multicolumn{2}{c|}{Conv2d} \\
        ~ & \multicolumn{2}{c|}{(repeat twice)} \\
        \cline{2-3}
        ~ & Conv2d($80 \times 80 \times 3$) & Conv2d($80 \times 80 \times 1$) \\
        ~ & Tanh & Tanh \\ \hline
        Output Shape & $80 \times 80 \times 3$ & $80 \times 80 \times 1$ \\ \hline
      \end{tabular}
    }
\end{table}

Table~\ref{tab:architecture_of_the_generator} shows the entire structure of $\mathcal{G}$ of generating the sticker $1$, and the other two branches is the same with it. And this network will sample the nosie from $N(0, 1)$ and map it to a sticker or a shape mask.

$\mathcal{D}$ also contains 3 the same sub-networks for separately distinguish the fake shapes produced by $\mathcal{G}$ at a time. The architecture of $\mathcal{D}$ is basically the flip of the structure of $\mathcal{G}$ for producing the shape masks, which is shown in Table~\ref{tab:architecture_of_the_discriminator}.

\begin{table}[t]
\caption{The architecture of $\mathcal{D}$}
\label{tab:architecture_of_the_discriminator}
\centering
\renewcommand\arraystretch{1.3}
\scalebox{0.80}{
    \begin{tabular}{|p{2.5cm}<{\centering}|p{5.5cm}<{\centering}|} \hline
    ~ & Network Structure \\ \hline
    Input Shape & $80 \times 80 \times 1$ \\ \hline
    \multirow{9}{*}{Discriminator} & Conv2d \\
    ~ & LayerNorm \\
    ~ & ReLU \\
    ~ & MaxPool2d \\
    ~ & (repeat twice) \\
    \cline{2-2}
    ~ & LayerNorm($5 \times 5 \times 640$) \\
    ~ & ReLU \\
    ~ & Reshape($1 \times 16000$) \\
    ~ & FC($16000 \times 1$)\\ \hline
    Output Shape & $1 \times 1$ \\ \hline
    \end{tabular}
}
\end{table}

\section{Guided Grad-CAM}\label{sec:guided_grad_cam}

Selvaraju et al.~\cite{selvaraju2017grad} proposed a novel approach, called as Guided Gradient-weighted Class Activation Mapping (Guided Grad-CAM), to utilize the gradients of any target class to produce the localization map highlighting the critical regions, which is widely used in constraining the perturbed areas when crafting the adversarial examples~\cite{deng2019generate,liu2019perceptual}. In this paper, the Guided Grad-CAM and its variants~\cite{chattopadhay2018grad} are introduced to localize stickers.

Guided Grad-CAM uses the gradient information flowing into the last conlutional layer of the feature extractor to analyze the importance of each region in images for a decision of interest. However, the original Guided Grad-CAM is designed for the networks whose activation funtion is the ReLU. Recently, some variants of ReLU (e.g., PReLU) have been widely applied in CNNs. The difference between PReLU and ReLU is the negative values are still meaningful when using PReLU activations. In this paper, we choose three FR systems, ArcFace~\cite{deng2019arcface}, CosFace~\cite{wang2018cosface}, FaceNet~\cite{schroff2015facenet}, as target models, two of which (i.e., CosFace, ArcFace) have PReLU activations. The details of their architectures can be found in Appendix~\ref{sec:architecture_of_face_recognition_system}. Inspired by the variant proposed by Chattopadhay et al.~\cite{chattopadhay2018grad}, the ReLU is used to clip out the negative gradients. Because the pixels with negative gradients are likely to belong to other classes in the image.

\section{3D Face Reconstruction}\label{sec:3d_face_reconstruction}

Tran et al.~\cite{tran2019towards} showed an application of 3D face reconstruction on adding stickers onto faces, which unwarps the facial texture into the UV space. The UV unwarping is the processing of unfolding the surface of 3D model to a 2D image that is called as the UV space, and the reverse process is refered to as the UV mapping. In the UV space, we can directly edit the texture by adding stickers. Finally, the edited texture images can be rendered using the estimated 3D face shape together with the parameters of the illumination model.

With the advancement of the 3D face reconstruction, some methods obtaining accurate 3D face shape from a single image have been proposed. Deng et al.~\cite{deng2019accurate} utilized a network, named as R-Net, to gain the 3D face shape, the texture image and the parameters of the illumination model (i.e., Spherical Harmonics that can be represented by $27$ parameters). When doing UV unwarping to generate texture images, this approach need UV coordinates to map the 3D vertices of face shape onto the 2D texture image, which is provided by Bas et al.~\cite{bas20173d}. Besides, the R-Net can estimate the parameters of Spherical Harmonics to approximate the scene luminance. After getting these information, the differentiable renderer can render the texture image according to the 3D face shape while fusing the reckoned ambient brightness.

The 3D face shape describes the human face shape and the texture corresponding to the input image, which consists of many 3D vertices (e.g., 36K in~\cite{deng2019accurate}, 39K in~\cite{tran2019towards}). The illumination model (e.g., Spherical Harmonics~\cite{ramamoorthi2001efficient} that can be represented by $27$ parameters) is used to approximate the scene light, and the camera model represents the matrix for the 3D-2D projection geometry. Comparing the original and rendered face images in Fig.~\ref{fig:workflow}, it is obvious that the estimated head pose and 3D face shape are all accurate and the brightness of left face on the rendered image is the same as the original image.

\section{The Architecture of FR Systems}\label{sec:architecture_of_face_recognition_system}

We will elaborately describe the architecture of FR systems, i.e., ArcFace, CosFace and FaceNet, which are illustrated in Table.~\ref{tab:architecture_of_face_recognition_systems}.

In this paper, the aim of attackers is not to deceive the face detector for not being detected but to implement the dodging and impersonating attacks. Thus, the attacked FR system should be the feature extractor and the MLP classifier. The feature extractor is already trained on a large-scale dataset (e.g., MS-Celeb-1M, VggFace2) for obtaining the best performance. And only the MLP classifier is trained on our small-scale dataset ($\rm{LFW^-}$ and VulFace).

\begin{table}[t]
\caption{The architecture of FR systems}
\label{tab:architecture_of_face_recognition_systems}
\centering
\renewcommand\arraystretch{1.3}
\scalebox{0.65}{
    \begin{tabular}{|c|c|c|c|} \hline
    ~ & ArcFace & CosFace & FaceNet \\ \hline
    Input Shape & $112 \times 112 \times 3$ & $112 \times 96 \times 3$ & $160 \times 160 \times 3$ \\ \hline
    Feature Extractor & LResNet34E-IR~\cite{deng2019arcface} & Sphere20~\cite{liu2017sphereface} & InceptionResNetV1~\cite{szegedy2016inception} \\ \hline
    \multirow{4}{*}{MLP Classifier} & \multicolumn{3}{c|}{FC($512 \times 256$)} \\
    ~ & \multicolumn{3}{c|}{Tanh} \\
    ~ & \multicolumn{3}{c|}{FC($256 \times 153$)} \\
    ~ & \multicolumn{3}{c|}{Softmax} \\ \hline
    Output Shape & \multicolumn{3}{c|}{$1 \times 153 $} \\ \hline
    \end{tabular}
}
\end{table}
The input dimension of three different FR systems are $112 \times 112 \times 3$, $112 \times 96 \times 3$ and $160 \times 160 \times 3$. And the MLP classifier consists of two linear layers and a tanh activation layer. FC refers to the fully connected layer. Since the number of identities in $\rm{LFW^-}$ and VulFace is 153, the output shape of the MLP classifier is $1 \times 153$.

Since our goal is the feature extractor and the MLP classifier, in our evaluations, we find the face detector cannot successfully detect human faces, which only happens in extreme cases.

\section{Algorithm For Training generator}\label{sec:training_generator}

Here is the algorithm for training the sticker generator.

The $2-9$ lines are applied to training $\mathcal{D}$. For improving the performance of $\mathcal{D}$, FaceAdv trains $\mathcal{D}$ $n_\mathcal{D}$ times when training $\mathcal{G}$ per time. During training $\mathcal{D}$, FaceAdv samples the noise $n$ and the real shape $s$ from shape templates.
Then, the sampled $n$ is fed into the generator $\mathcal{G}$ to obtain fake shapes.
The crafted shapes and real shapes are input into the discriminator $\mathcal{D}$ to calculate the $\mathcal{L}_\mathcal{D}$.
Finally, FaceAdv utilize the Adam optimizer to update parameters of $\mathcal{D}$.

The line $10-20$ are used to training $\mathcal{G}$. FaceAdv calculates $\mathcal{L}^s_{\mathcal{G}}$, $\mathcal{L}_{tv}$ and $\mathcal{L}_{adv}$ in turn, which makes up the total loss of $\mathcal{G}$.
When obtaining $\mathcal{L}_{adv}$, face images of attackers and crafted stickers should be fed into the transformer $\mathcal{T}$ and then FaceAdv chooses different equations accroding to the type of attacks.
And, finally, the total loss $\mathcal{L}_{\mathcal{G}}$ is employed to optimize parameters of $\mathcal{G}$.

\begin{algorithm}[t!]
    \small
    \caption{Training the generator in FaceAdv}
    \label{algorithm:faceadv_training}
    \begin{algorithmic}[1]
        \Require Attacker images $X_A$, auxiliary dataset $A$, shape dataset $S$, noise distribution $N$, batch size $m$, training epochs $epochs$, generator $\mathcal{G}$, discriminator $\mathcal{D}$, target FR $\mathcal{F}$
        \Ensure The generator $\mathcal{G}$
        \For{$epoch=1 \to epochs$}
        \Statex \ \ \ \ // train $\mathcal{D}$
        \For {$t=1 \to n_{\mathcal{D}}$}
            \State Sample $\{n^{(i)}\}_{i=1}^m \sim \mathbb{P}_n$ a batch from $N$
            \State Sample $\{s^{(i)}\}_{i=1}^m \sim \mathbb{P}_s$ a batch from $S$
            \State Sample $\{\epsilon^{(i)}\}_{i=1}^m \sim U[0, 1]$
            \State $\hat{s} \gets \epsilon s + (1 - \epsilon)\mathcal{G}(n)$
            \State $\mathcal{L}_{\mathcal{D}} \gets \mathcal{D}(\mathcal{G}(n)) - \mathcal{D}(s) + \lambda \lbrack (\Vert \bigtriangledown_{\hat{s}} \mathcal{D}(\hat{s}) \Vert_2 - 1) ^2 \rbrack$
            \State $\mathcal{D} \gets Adam(\bigtriangledown_{\mathcal{D}} \mathcal{L}_{\mathcal{D}}, \mathcal{D}, r)$
        \EndFor
        \Statex \ \ \ \ // train $\mathcal{G}$
        \State $\mathcal{L}^s_{\mathcal{G}} \gets - \mathcal{D}(\mathcal{G}(n))$
        \State $\mathcal{L}_{tv} \gets \mathcal{L}_{tv}(n)$
        \State Sample $\{x_A^{(i)}\}_{i=1}^m \sim \mathbb{P}_{x_A}$ a batch from $X_A$
        \State Sample $\{a^{(i)}\}_{i=1}^m \sim \mathbb{P}_{a}$ a batch from $A$
        \If {$dodging$}
            \State $\mathcal{L}_{adv} \gets -\mathcal{F}(\mathcal{T}(x_A, \mathcal{G}(n), a), P_A)$
        \Else
            \State $\mathcal{L}_{adv} \gets \mathcal{F}(\mathcal{T}(x_A, \mathcal{G}(n), a), P_B)$
        \EndIf
        \State $\mathcal{L}_{\mathcal{G}} \gets \mathcal{L}_{\mathcal{G}}^s + \alpha \mathcal{L}_{adv} + \beta \mathcal{L}_{tv}$
        \State $\mathcal{G} \gets Adam(\bigtriangledown_{\mathcal{G}} \mathcal{L}_{\mathcal{G}}, \mathcal{G}, r)$
        \EndFor
    \end{algorithmic}
\end{algorithm}

\section{Experimental Site}\label{sec:experimental_site}

In this section, we will elaborately describe the details about how experimental devices are placed.

\begin{figure}[t!]
    \centering
    \subfigure[The front view]{
        \begin{minipage}[b]{0.22\textwidth}
        \centering
        \includegraphics[width=1\textwidth]{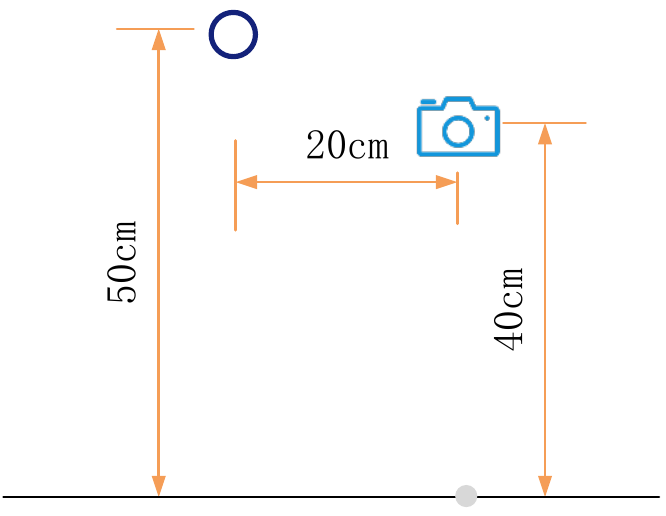}
        \end{minipage}
    }
    \subfigure[The top view]{
        \begin{minipage}[b]{0.22\textwidth}
        \centering
        \includegraphics[width=1\textwidth]{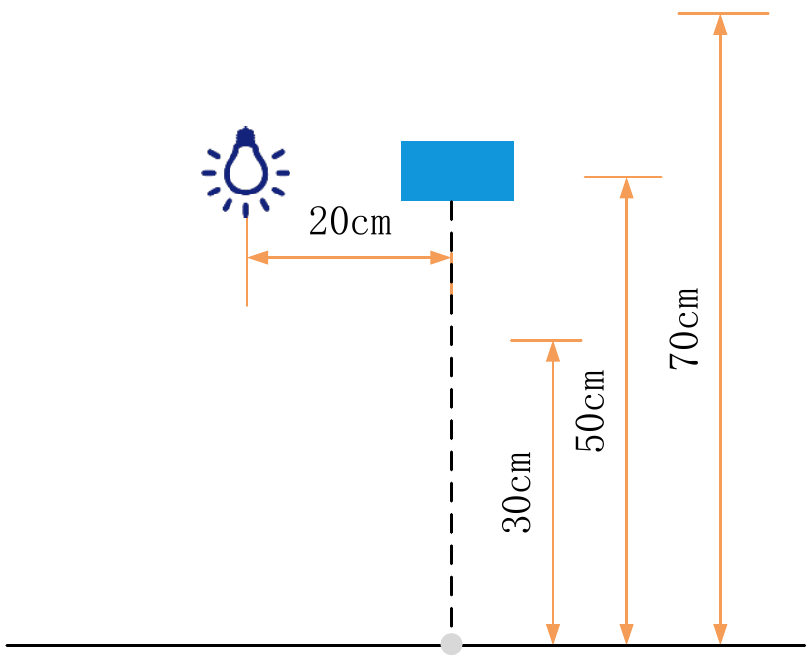}
        \end{minipage}
    }
    \caption{The front view and top view of experimental site. The blue circle represents the light and the rectangular means the camrea. In addition, the grey dot is the position where the head of volunteers shoud be.}
    \label{fig:experimental_site}
\end{figure}

Fig.~\ref{fig:experimental_site} shows the positions of the camera and light, which are all placed on a large table. And volunteers are set on the place where the grey dot is in Fig.~\ref{fig:experimental_site}. Since behind the camera and the light is a wall, when conducting the effect of the head pose on the performance of FaceAdv, we paste green dots on positions of corresponding angles.

\section{Attacking VGGFace}\label{sec:attacking_vggface}

\begin{table}[t]
    \caption{Success rate of attacking VGGFace}
    \label{tab:success_rate_of_attacking_VGGFace}
    \centering
    \renewcommand\arraystretch{1.3}
    \scalebox{0.76}{
        \begin{tabular}{|c|c|c|c|} \hline

        Scenario & Mode & Method & VGGFace \\ \hline
        \multirow{4}{*}{Digital} & \multirow{2}{*}{Dodging} & FaceAdv & $100.00\%$ \\
        \cline{3-4}
        ~ & ~ & AGNs & $100.00\%$ \\
        \cline{2-4}
        ~ & \multirow{2}{*}{Impersonating} & FaceAdv & $100.00\%$ \\
        \cline{3-4}
        ~ & ~ & AGNs & $79.44\%$  \\ \hline
        \multirow{4}{*}{Physical} & \multirow{2}{*}{Dodging} & FaceAdv & $100.00\%$  \\
        \cline{3-4}
        ~ & ~ & AGNs & $100.00\%$  \\
        \cline{2-4}
        ~ & \multirow{2}{*}{Impersonating} & FaceAdv & $33.33\%$ \\
        \cline{3-4}
        ~ & ~ & AGNs & $11.20\%$  \\ \hline
        \end{tabular}
    }
\end{table}

We conduct a relatively smaller experiment to prove the effectiveness of FaceAdv when attacking the VGGFace. The feature extractor of VGGFace takes a $244\rm{px} \times 244\rm{px}$ aligned face image as input and produces a vector representation of the face that has 4096 dimensions. And the MLP classifier of the VGGFace is the same as that of ArcFace, CosFace and FaceNet. The VGGFace separately achieves $97.27\%$, $99.33\%$ and $100.00\%$ success rate in LFW, $\rm{LFW^-}$ and VulFace.

We randomly select 3 volunteers from VulFace as attackers and 3 victims from the $\rm{LFW^-}$ for each attacker for implementing dodging and impersonating attacks. Therefore, there are $3 \times 3 \times 2 + 3 \times 2 = 24$ experiments to be carried out. Table~\ref{tab:success_rate_of_attacking_VGGFace} shows results of attacking the VGGFace. we can find FaceAdv and AGNs can also successfully attack VGGFace in physical and digital scenarios. However, the success rate of FaceAdv is relatively higher than that of AGNs.

\end{document}